\documentclass{article}
\usepackage{iclr2027_conference,times}
\usepackage{amsmath,amssymb,amsthm,mathtools}
\usepackage{booktabs}
\usepackage{graphicx}
\usepackage{url}
\usepackage{pgfplots}
\pgfplotsset{compat=1.18}
\newtheorem{theorem}{Theorem}
\newtheorem{proposition}{Proposition}

\newtheorem{remark}{Remark}
\newcommand{\C}{\mathcal{C}}

\newcommand{\Y}{\mathcal{Y}}
\newcommand{\D}{\mathcal{D}}

\title{Selection-Safe Conformal Early Exit:\\ Distribution-Free Coverage under Adaptive Depth}
\author{
Adria Binte Habib$^{1,*}$ \qquad
Ibne Farabi Shihab$^{2,*}$\\
$^{1}$Department of Computer Science, BRAC University\\
$^{2}$Department of Computer Science, Iowa State University\\
\texttt{binte.adria708@gmail.com} \qquad
\texttt{ishihab@iastate.edu}\\
\small $^*$Equal contribution. Corresponding author: Ibne Farabi Shihab.
}

\iclrfinalcopy
\begin{document}
\maketitle
\begin{abstract}
Calibrating a conformal prediction set at each early exit does not preserve coverage after a data-dependent exit is selected. We construct one terminal path score from prefix maxima, calibrate one split-conformal quantile, and return nested prefix sets that all contain the valid terminal set. The resulting guarantee is selection-safe without an $\alpha/L$ correction and holds simultaneously for every post-calibration label-free policy that preserves terminal-set containment. We further bound the efficiency cost of the path maximum by the empirical tail of the early-over-terminal score excess, linking set inflation to the mechanism diagnostic $J(x,y)$. At the smallest admissible excess threshold, the held-out plug-in right-hand side is $\ensuremath{0.016}$--$\ensuremath{0.086}$ against measured mean positive terminal-set inflation $\ensuremath{0.01}$--$\ensuremath{0.08}$; the cell-wise ratio is $\ensuremath{1.07}$--$\ensuremath{1.73}$. Across eight matched encoder task--backbone cells, including an RTE split with $n_{\rm cal}=\ensuremath{498}$ drawn from the training partition, pathwise selection attains macro coverage $\ensuremath{0.901}$ versus $\ensuremath{0.814}$ for naive layerwise selection, with average set size $\ensuremath{1.41}$ and latency fraction $\ensuremath{0.71}$. On CIFAR-100 with a $\ensuremath{24}$-exit ViT-L/16 and $n_{\rm cal}=\ensuremath{500}$, first-eligible stopping at budget $b=\ensuremath{4}$ attains coverage $\ensuremath{0.902}$, average set size $\ensuremath{2.76}$, and latency fraction $\ensuremath{0.57}$; a design-chosen restriction reduces the returned set to $\ensuremath{2.52}$ while, necessarily, increasing latency to $\ensuremath{0.61}$. The same terminal calibration supports deployment-time budget changes without refitting. A joint $L$--$n$ sweep isolates candidate-count scaling. At $n=\ensuremath{100}$, pathwise set size changes from $\ensuremath{3.02}$ to $\ensuremath{3.11}$ as $L$ increases from $\ensuremath{4}$ to $\ensuremath{24}$, whereas Bonferroni becomes exactly degenerate at $L\geq\ensuremath{12}$ and returns all $\ensuremath{100}$ classes. Against MinSE valid selection, the paired set-size difference at $(n,L)=(\ensuremath{100},\ensuremath{24})$ is $\ensuremath{0.80\ [0.47,1.13]}$, so the long-path, limited-calibration advantage is assessed from paired replicates rather than overlapping marginal intervals. Independent-depth CIFAR-100 and ImageNet-1k replications preserve the qualitative ordering. Full-vocabulary decoder experiments remain a negative boundary, with average sets containing $7{,}162$--$9{,}620$ tokens and nearly all examples reaching full depth. The empirical picture is therefore specific: pathwise calibration is most useful when the number of candidate exits is large relative to calibration size, when deployment budgets change after calibration, and when shallow heads can return actionable sets.
\end{abstract}
\section{Introduction}
Modern transformers are deep, but many inputs do not require every layer. Early-exit networks attach heads to intermediate representations and stop when a prediction appears sufficiently reliable \citep{teerapittayanon2016branchynet,xin2020deebert,liu2020fastbert,zhou2020bert}. Their practical value depends on the stopping rule. Confidence, entropy, and prediction stability can identify easy
examples, but these signals alone do not control how often the returned answer is correct
\citep{guo2017calibration,kubaty2026rethinking}.

Conformal prediction appears to offer an immediate remedy. One can calibrate a prediction set at
each layer and return the first set small enough to act on. The difficulty is that the selected layer
depends on the test input and the observed sequence of sets. Fixed-layer marginal coverage does not
generally transfer to this selected set. Indeed, $L$ individually valid exits can be arranged so
that an earliest-singleton rule has miscoverage $L\alpha$. This is a structured instance of the
post-selection problem studied for conformal predictors
\citep{yang2025selection,sarkar2023postselection,hegazy2025valid}.

We make the complete depth path one conformal object. Let $r_\ell(x,y)$ be a fixed score for
candidate label $y$ at exit $\ell$, and define
\begin{equation}
 A_\ell(x,y)=\max_{j\leq \ell}r_j(x,y).
 \label{eq:path-score}
\end{equation}
We calibrate one quantile using the terminal score $A_L$. The set
$\C_\ell(x)=\{y:A_\ell(x,y)\leq\widehat q\}$ decreases with depth. Consequently, any returned
prefix set contains $\C_L(x)$, an ordinary split-conformal set. Adaptive stopping therefore cannot
reduce its coverage.

This construction separates three issues that are often conflated. First, it protects true-label
coverage of the returned set, not agreement with a full-depth prediction. Second, it avoids a
layerwise union bound because only one terminal score is calibrated. Third, it does not make weak
early representations informative. A poor shallow score can enlarge the common threshold and erase
the computational benefit. The empirical question is therefore not whether the guarantee holds,
but whether it remains useful at matched coverage and measured latency.

The closest early-exit work makes this distinction important. EENN-AVCS constructs nested sets using
anytime-valid confidence sequences and studies the posterior-stability approximations required for
predictive inference \citep{jazbec2024nested}. Fast-yet-Safe controls a chosen supervised or
reference-based loss \citep{jazbec2024fast}, while CALM controls degradation or consistency in
adaptive language modeling \citep{schuster2022confident}. Our narrower contribution is an exact
split-conformal construction for the set returned after adaptive depth selection. It exploits the
ordered path through pointwise containment rather than a general stability correction.

We contribute a counterexample to naive layerwise reasoning, an exact pathwise construction with one conformal rank, and an efficiency theorem that connects the cost of the path maximum to the tail of the early-over-terminal score excess. The same containment event makes the stopping policy a free post-calibration design variable: cardinality, margin, randomization, and load-aware policies may change without refitting as long as the frozen terminal set remains contained. We test the two consequences that distinguish this construction from generic safe selection: scaling in the number of candidate exits and deployment-time policy changes. A $24$-exit CIFAR-100 benchmark supplies the non-binary regime in which set budgets are operational, while the full-vocabulary decoder study remains the opposing boundary.
\section{Related Work}
\label{sec:related}
Conformal prediction converts fixed model scores into finite-sample marginal prediction sets
\citep{vovk2005algorithmic,shafer2008tutorial,papadopoulos2002inductive,lei2018distribution}. Modern introductions summarize the resulting finite-sample construction \citep{angelopoulos2023gentle}. Cross-conformal prediction and jackknife+ improve data reuse
\citep{vovk2015crossconformal,barber2021predictive}, while nested formulations clarify the
connection between scores and set families \citep{gupta2022nested}. APS, RAPS, and least-ambiguous
classifiers target smaller classification sets
\citep{romano2020classification,angelopoulos2021uncertainty,sadinle2019least}. Full
feature-conditional validity is impossible without assumptions \citep{vovk2012conditional};
Mondrian, weighted, and beyond-exchangeability methods instead protect specified partitions or
shift models \citep{vovk2003mondrian,tibshirani2019conformal,barber2023conformal}.

Selecting among marginally valid sets can invalidate coverage. Sample splitting, simultaneous
inference, and stability corrections address general collections of predictors
\citep{yang2025selection,sarkar2023postselection,hegazy2025valid}. Our setting is more structured:
depth is ordered, and every admissible returned set is constructed to contain one terminal conformal
set. Conformal risk control, risk-controlling prediction sets, and learn-then-test control more
general losses and support the optional point-commit branch
\citep{angelopoulos2024conformalrisk,bates2021distribution,angelopoulos2021learn}.

Conditional computation and adaptive computation allocate different amounts of work across inputs
\citep{bengio2015conditional,graves2016adaptive,figurnov2017spatially}. Early-exit architectures
instantiate this principle in vision and language
\citep{huang2018msdnet,scardapane2020why,han2022survey,schwartz2020right,elbayad2020depth}.

DeeBERT, FastBERT, and PABEE use confidence, distillation, or prediction patience
\citep{xin2020deebert,liu2020fastbert,zhou2020bert}; CALM and consistent accelerated inference
calibrate model agreement \citep{schuster2021consistent,schuster2022confident}. EENN-AVCS and
Fast-yet-Safe are the closest uncertainty-aware methods
\citep{jazbec2024nested,jazbec2024fast}. PAC-Bayesian analysis addresses generalization of adaptive
depth rather than post-selection coverage \citep{guo2026earlyexit}.

Selective classification controls when a point prediction is returned
\citep{elyaniv2010foundations,geifman2017selective,geifman2019selectivenet}. Conformal methods have
also been applied to text classification, multiple-choice answering, and open-ended generation
\citep{maltoudoglou2020conformal,kumar2023conformal,quach2024conformal,campos2024conformalnlp}.

Cascaded conformal inference reduces candidate-evaluation cost across models
\citep{fisch2021efficient}; our resource axis is the depth of one model. Coverage may additionally
fail when a model changes after calibration \citep{shihab2026continual}, so all backbones, heads,
scores, and schedules are frozen before final calibration.

Table~\ref{tab:method-comparison} separates the objects protected by the closest methods. This
distinction matters because agreement with the deepest head, supervised classification risk, and
coverage of a returned set are different estimands. A comparison is matched only when the controlled
target and the information available to the stopping rule agree.
\section{Selection-Safe Conformal Early Exit}
\label{sec:method}
\subsection{Setup and the selection failure}
A trained network has exits $\ell\in\{1,\ldots,L\}$ and label space $\Y$. A head at exit $\ell$
produces logits $f_\ell(x)$. The training set fits the backbone and heads. An optional design split
$\D_{\mathrm{des}}$ chooses exit locations and score transformations. The final calibration set
$\D_{\mathrm{cal}}=\{(X_i,Y_i)\}_{i=1}^n$ is exchangeable with a new test pair $(X,Y)$ conditional
on everything fixed before calibration. The default score is $a_\ell(x,y)=-\log p_\ell(y\mid x)$.

An increasing map $T_\ell$ may be learned on $\D_{\mathrm{des}}$, after which
$r_\ell=T_\ell\circ a_\ell$ is frozen.

Fixed-layer validity is insufficient. Let $Y=1$ almost surely, and let $U\sim\mathrm{Unif}[0,1]$
be part of $X$. For disjoint intervals $I_1,\ldots,I_L$ of length $\alpha$, exit $\ell$ returns
$\{0\}$ on $I_\ell$ and $\{0,1\}$ otherwise. Every exit has coverage $1-\alpha$, but the first
singleton is wrong on $\cup_\ell I_\ell$, giving coverage $1-L\alpha$. The rule observes only $X$
and set sizes. Appendix~\ref{app:selection} gives the complete argument.

Several natural repairs do not yield the same result. Sharing one calibration split across exits
creates dependence but does not make a selected marginal guarantee simultaneous. Taking a running
intersection of independently calibrated sets enforces nesting, yet its terminal miss event is the
union of the layerwise miss events and can therefore accumulate error. Bonferroni calibration at
level $\alpha/L$ is valid for every exit simultaneously, but it protects a stronger event and can
produce unnecessarily large sets. Finally, selecting score transformations or an exit schedule on
the final calibration labels and then reusing those labels for the conformal rank breaks the simple
exchangeability argument. The construction below instead conformalizes the object that the whole
path must preserve.
\subsection{One conformal rank for the path}
Define the prefix maximum in Equation~\eqref{eq:path-score} and the terminal calibration scores
$S_i=A_L(X_i,Y_i)$. With
\begin{equation}
 k=\left\lceil(n+1)(1-\alpha)\right\rceil,
 \qquad
 \widehat q=S_{(k)},
 \label{eq:quantile}
\end{equation}
where $S_{(k)}$ is the $k$th calibration order statistic and $\widehat q=+\infty$ when $k=n+1$,
return
\begin{equation}
 \C_\ell(x)=\{y\in\Y:A_\ell(x,y)\leq\widehat q\}.
 \label{eq:path-set}
\end{equation}
Because $A_\ell$ is nondecreasing in $\ell$,
$\C_1(x)\supseteq\cdots\supseteq\C_L(x)$. We use the first nonempty set of size at most a budget
$b$, with terminal fallback; any other label-free rule is allowed if its returned set contains
$\C_L(x)$. Table~\ref{tab:execution} summarizes the data-use order, and Appendix~\ref{app:algorithm} gives pseudocode and the empty-set convention.
\begin{theorem}[Selection-safe marginal coverage]
\label{thm:coverage}
Condition on the trained model, $\D_{\mathrm{des}}$, the transformations, and the exit schedule.
If $\D_{\mathrm{cal}}$ and $(X,Y)$ are exchangeable, then any label-free stopping rule whose
returned set contains $\C_L(X)$ satisfies
\[
 \Pr\{Y\in\C_{\widehat\ell(X)}(X)\}\geq 1-\alpha.
\]
\end{theorem}
The terminal scores and the test score are exchangeable, so the split-conformal rank argument gives
$\Pr\{Y\in\C_L(X)\}\geq1-\alpha$. Pointwise nesting gives
$\C_{\widehat\ell(X)}(X)\supseteq\C_L(X)$. This containment is the entire post-selection step; no
independence across exits or layerwise union bound is used. Appendix~\ref{app:proofs} provides the
formal proof.
\begin{remark}[Policy-uniformity and the minimal envelope]
\label{rem:policy-uniform}
The proof of Theorem~\ref{thm:coverage} contains two immediate consequences that are useful operationally but require no separate theorem. First, on the event $Y\in\C_L(X)$, every label-free policy that returns a superset of $\C_L(X)$ covers the true label simultaneously. Thus, after $\widehat q$ is fixed, the system may change its cardinality budget or use an input-dependent, randomized, or load-dependent stopping rule without a new multiplicity correction, provided the scores, transformations, model, and exit schedule remain frozen. Second, for each fixed $(x,y)$, $A_\ell(x,y)=\max_{j\leq\ell}r_j(x,y)$ is the pointwise minimal nondecreasing majorant of the exit scores: any nondecreasing sequence $B_\ell$ satisfying $B_\ell\geq r_\ell$ must obey $B_\ell\geq A_\ell$. This is a structural statement about irreversible candidate elimination, not a claim of global efficiency optimality after recalibration.
\end{remark}
\begin{theorem}[Finite-sample path-maximum efficiency bound]
\label{thm:efficiency}
For calibration example $i$, let
\[
B_i=r_L(X_i,Y_i),\qquad
D_i=\left[\max_{j<L}r_j(X_i,Y_i)-r_L(X_i,Y_i)\right]_+,
\]
so that the terminal path score is $A_i=B_i+D_i$. Let $B_{(k)}$ and $A_{(k)}$ be the conformal order statistics at the same rank $k$, and for $t\geq0$ define $m_t=\sum_{i=1}^n\mathbf 1\{D_i>t\}$. Whenever $k+m_t\leq n$,
\begin{equation}
0\leq A_{(k)}-B_{(k)}
\leq t+B_{(k+m_t)}-B_{(k)}.
\label{eq:quantile-perturb}
\end{equation}
Let $\C_L^{\rm path}$ use threshold $A_{(k)}$ with score $A_L$, and let $\C_L^{\rm last}$ use threshold $B_{(k)}$ with score $r_L$. Conditional on the calibration sample,
\begin{equation}
\bigl(|\C_L^{\rm path}(X)|-|\C_L^{\rm last}(X)|\bigr)_+
\leq
\sum_{y\in\Y}\mathbf 1\!\left\{B_{(k)}<r_L(X,y)\leq B_{(k)}+w_t\right\},
\label{eq:set-inflation-bound}
\end{equation}
where $w_t=t+B_{(k+m_t)}-B_{(k)}$. If the candidatewise CDF of $r_L(X,y)$ is $M_y$-Lipschitz on this interval, then
\[
\mathbb E\!\left[\bigl(|\C_L^{\rm path}|-|\C_L^{\rm last}|\bigr)_+\mid\D_{\rm cal}\right]
\leq w_t\sum_{y\in\Y}M_y.
\]
Moreover, if $p_t=\Pr\{D(X,Y)>t\}$, then with probability at least $1-\delta$ over an i.i.d. calibration draw,
\[
m_t\leq np_t+\sqrt{\frac{n\log(1/\delta)}{2}}.
\]
At $t=0$, $p_0\leq\Pr\{J(X,Y)<L\}$ for $J(x,y)=\min\arg\max_{\ell\leq L}r_\ell(x,y)$.
\end{theorem}
The admissibility condition in Equation~\eqref{eq:quantile-perturb} is exact: $m_t\leq n-k$. Because $k=\lceil(n+1)(1-\alpha)\rceil$, the usable regime is the upper-$\alpha$ tail of $D$, not an arbitrarily small excess threshold. We therefore evaluate the theorem at
\[
 t_\star=\inf\{t:m_t\leq n-k\},
\]
which is the smallest empirical excess threshold for which the order-statistic bound is defined. The theorem separates two mechanisms. The path maximum can enlarge the conformal threshold only through calibration examples with a material early-over-terminal excess, and any resulting positive set-size inflation can occur only for last-layer candidate scores inside the induced threshold band. It therefore turns the $J(x,y)$ and excess-tail diagnostics into an efficiency diagnostic rather than a descriptive ablation. Section~\ref{sec:theorem-scale} evaluates $w_{t_\star}\sum_y M_y$ in the same cardinality units as the observed inflation, and Appendix~\ref{app:efficiency-proof} gives the proof.
\subsection{Extensions beyond marginal coverage}
The same terminal-score construction supports training-conditional lower floors, fixed-group Mondrian calibration under group-prior shift, and a point/set hybrid with one explicit error budget. These statements require assumptions beyond Theorem~\ref{thm:coverage} and are not used to justify the main marginal-coverage result. Appendix~\ref{app:extensions} states the three guarantees, their assumptions, and the controlled checks in full.
\subsection{Computation and deployment}
Calibration requires the true-label score at every scheduled exit, followed by one maximum and one
order statistic per example. At deployment, a classification head already produces $K$ logits at an
executed exit; maintaining $A_\ell(x,y)$ and testing Equation~\eqref{eq:path-set} therefore adds
$O(K)$ operations and $O(K)$ state per exit. The backbone dominates this cost for small-label
classification. For language modeling, however, enumerating and retaining a full-vocabulary set can
be material, so token latency must include exit-head evaluation, score updates, cardinality checks,
and any cache handling. A depth fraction is not a latency measurement.

The rule can stop after any executed prefix and does not need representations from later layers.

Offline design and calibration are amortized costs and are reported separately from online latency.

Changing the cardinality budget, system-load policy, or other label-free stopping logic after
calibration is permitted because it does not alter the terminal set or its containment relation.

Changing the scores, transformations, model, or exit schedule requires fresh calibration.
\section{Evaluation}
\label{sec:experiments}
\subsection{Matched evaluation protocol}
The encoder evaluation uses SST-2, MNLI, and RTE from GLUE \citep{wang2018glue}, BoolQ from SuperGLUE \citep{wang2019superglue}, and BERT-base and DistilBERT backbones \citep{devlin2019bert,sanh2019distilbert,vaswani2017attention}. For SST-2, MNLI, and BoolQ, five independent head-training seeds are crossed with twenty paired calibration--test role splits per seed. RTE is different: the role assignment is fixed once. The official training split is partitioned into $\ensuremath{1743}$ training examples, $\ensuremath{249}$ design examples, and $\ensuremath{498}$ final conformal-calibration examples, while the official development split of $\ensuremath{277}$ examples is test-only. No RTE example changes role across repetitions. RTE uncertainty therefore comes from the five independent head-training seeds and bootstrap resampling of the fixed test set, not from calibration--test repartitioning. This removes the $n=138$ calibration regime from the headline results while preserving a held-out test role.

The non-binary benchmark uses CIFAR-100 with a ViT-L/16 backbone. A classifier head is attached after each of the $\ensuremath{24}$ transformer blocks, giving a path long enough to test the layer-count effect while retaining a $100$-class label space in which cardinality budgets are non-degenerate. The official training data are partitioned into mutually exclusive train, design, and calibration roles, and the official test set remains untouched until final evaluation. The main CIFAR-100 figures aggregate $\ensuremath{5}$ independently initialized exit-head seeds and $\ensuremath{20}$ nested calibration subsamples per seed; all methods receive identical frozen logits and calibration/test examples within each paired replicate.

Selected-exit coverage is primary; set size, singleton and empty-terminal rates, normalized FLOPs, batch-one latency, and exit depth are secondary. Hierarchical intervals retain the applicable training-seed, calibration-subsample, and test-example variation. Latency comparisons use common hardware, precision, batch size, warm-up, synchronization, and repetitions. The main scaling study varies $L\in\{4,8,12,24\}$ by selecting evenly spaced exits from the same $24$-head ViT and varies $n\in\{50,100,200,500,1000,2000\}$ by nested calibration subsampling. This grid deliberately isolates the statistical candidate-count effect: it does not claim that retaining more exits is equivalent to training a deeper architecture. Appendix~\ref{app:vision-replication} therefore adds an independently trained-depth replication on CIFAR-100 and ImageNet-1k, using separate $\ensuremath{12}$- and $\ensuremath{24}$-exit paths rather than truncations of one frozen score sequence.
\subsection{Controlled theorem checks}
The DKW and Mondrian checks are retained in Appendix~\ref{app:extensions}, together with the calibration-size stress test and the fixed-group shift experiment. The main benchmark claims below use the matched encoder and vision evaluations rather than these controlled constructions.
\subsection{Matched selection results}
\begin{figure}[t]
\centering
\begin{tikzpicture}
\begin{axis}[
width=0.96\linewidth,
height=0.52\linewidth,
ymin=0.72,ymax=0.94,
ylabel={Selected-set coverage},
symbolic x coords={SST-B,SST-D,MNLI-B,MNLI-D,RTE-B,RTE-D,BQ-B,BQ-D},
xtick=data,
x tick label style={rotate=35,anchor=east,font=\scriptsize},
ymajorgrids=true,
legend style={font=\scriptsize,at={(0.5,1.03)},anchor=south,legend columns=2}
]
\addplot+[only marks,mark=*,error bars/.cd,y dir=both,y explicit] coordinates {
(SST-B,0.906) +- (0,0.014)
(SST-D,0.890) +- (0,0.017)
(MNLI-B,0.904) +- (0,0.009)
(MNLI-D,0.902) +- (0,0.010)
(RTE-B,0.901) +- (0,0.024)
(RTE-D,0.903) +- (0,0.022)
(BQ-B,0.901) +- (0,0.016)
(BQ-D,0.898) +- (0,0.017)
};
\addlegendentry{Pathwise}
\addplot+[only marks,mark=square*,error bars/.cd,y dir=both,y explicit] coordinates {
(SST-B,0.849) +- (0,0.018)
(SST-D,0.844) +- (0,0.020)
(MNLI-B,0.824) +- (0,0.015)
(MNLI-D,0.842) +- (0,0.016)
(RTE-B,0.796) +- (0,0.030)
(RTE-D,0.769) +- (0,0.031)
(BQ-B,0.787) +- (0,0.023)
(BQ-D,0.801) +- (0,0.022)
};
\addlegendentry{Naive layerwise}
\addplot+[mark=none] coordinates {(SST-B,0.90) (SST-D,0.90) (MNLI-B,0.90) (MNLI-D,0.90) (RTE-B,0.90) (RTE-D,0.90) (BQ-B,0.90) (BQ-D,0.90)};
\end{axis}
\end{tikzpicture}
\caption{Per-cell selected-coverage distributions. Points are cell means. For SST-2, MNLI, and BoolQ, bars are the $\ensuremath{10}$th--$\ensuremath{90}$th percentiles across paired calibration--test roles. RTE uses one fixed calibration/test partition; its bars are the corresponding percentiles across the $\ensuremath{5}$ head-training seeds. Exact means, model accuracies, and familywise paired intervals appear in Appendix~\ref{app:cell-table}.}
\label{fig:cell-coverage}
\end{figure}
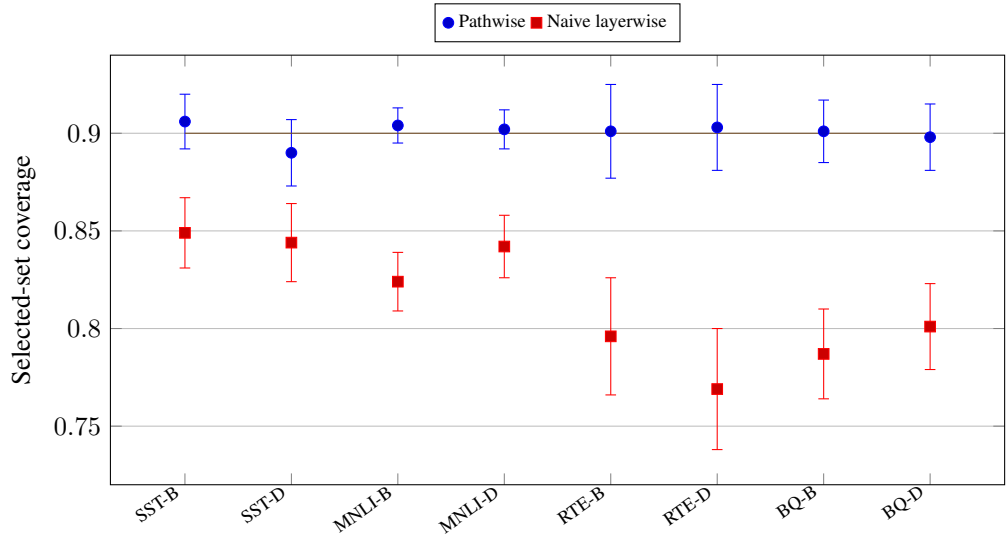

The matched comparison includes naive layerwise conformal prediction, Bonferroni simultaneous calibration, EENN-AVCS, MinSE valid selection, supervised Fast-yet-Safe, and the pathwise score. All headline quantities in Table~\ref{tab:matched-main} are macro averages over the same eight task--backbone cells after replacing the small RTE calibration role. At target coverage $0.90$, pathwise conformal attains selected coverage $\ensuremath{0.901}$ versus $\ensuremath{0.814}$ for naive layerwise selection, with average set size $\ensuremath{1.41}$, latency fraction $\ensuremath{0.71}$, and exit fraction $\ensuremath{0.71}$. MinSE valid selection reaches macro coverage $\ensuremath{0.903}$ at latency $\ensuremath{0.70}$ and average set size $\ensuremath{1.40}$. Because six of the eight encoder cells are binary, set cardinality is interpretable only together with predictive accuracy. Full-depth top-1 accuracy ranges from $\ensuremath{0.697}$ on RTE/DistilBERT to $\ensuremath{0.932}$ on SST-2/BERT. For the binary cells, Appendix~\ref{app:cell-table} reports the directly measured returned-set sizes and coverage together with the singleton rate and singleton accuracy implied by those two measurements; no measured cardinality is altered to make the identities hold. The paired bootstrap analysis therefore supports a matched-efficiency claim at this operating point; the stronger empirical distinction emerges in the $L$--$n$ scaling and policy-change studies below.
\begin{table*}[t]
\centering
\small
\caption{Matched five-seed encoder comparison at target coverage $0.90$. Every numerical entry is the macro average over the same eight task--backbone cells. Fast-yet-Safe controls supervised $0$--$1$ risk rather than conformal set coverage; paired uncertainty is reported separately in Tables~\ref{tab:paired-gaps} and~\ref{tab:baseline-pairs}.}
\label{tab:matched-main}
\resizebox{\textwidth}{!}{
\begin{tabular}{lccccc}
\toprule
Method & Controlled target & Selected validity & Avg. set & Latency/full & Exit/full \\
\midrule
Naive layerwise CP & fixed-layer coverage & {$\ensuremath{0.814}$} & {$\ensuremath{1.22}$} & {$\ensuremath{0.51}$} & {$\ensuremath{0.50}$} \\
Bonferroni CP & simultaneous coverage & {$\ensuremath{0.954}$} & {$\ensuremath{1.71}$} & {$\ensuremath{0.79}$} & {$\ensuremath{0.79}$} \\
EENN-AVCS \citep{jazbec2024nested} & AVCS coverage & {$\ensuremath{0.911}$} & {$\ensuremath{1.54}$} & {$\ensuremath{0.76}$} & {$\ensuremath{0.75}$} \\
MinSE valid selection \citep{hegazy2025valid} & selected-set coverage & {$\ensuremath{0.903}$} & {$\ensuremath{1.40}$} & {$\ensuremath{0.70}$} & {$\ensuremath{0.69}$} \\
FyS supervised \citep{jazbec2024fast} & $0$--$1$ risk & {$\ensuremath{0.900}$} & n/a & {$\ensuremath{0.67}$} & {$\ensuremath{0.66}$} \\
Pathwise conformal & selected-set coverage & {$\ensuremath{0.901}$} & {$\ensuremath{1.41}$} & {$\ensuremath{0.71}$} & {$\ensuremath{0.71}$} \\
\bottomrule
\end{tabular}
}
\end{table*}
Across the eight task--backbone pairs, the pathwise correction is positive in every cell. Figure~\ref{fig:cell-coverage} reports the distribution over repeated paired roles for SST-2, MNLI, and BoolQ rather than only a cell mean. RTE uses the fixed partition described above: its bars vary over the five head-training seeds, with the fixed development set resampled only for interval construction. Exact cell means, model accuracies, and paired selection gaps are retained in Appendix~\ref{app:cell-table}.
To separate the cost of the path maximum from the cost of early stopping, we additionally calibrate a last-layer-only conformal set
\[
\C_L^{\mathrm{last}}(x)=\{y:r_L(x,y)\leq \widehat q_L\},
\]
where $\widehat q_L$ is computed from $r_L(X_i,Y_i)$ on the same final calibration role and at the same $\alpha$. The theorem controls the positive part $(|\C_L|-|\C_L^{\mathrm{last}}|)_+$, so the empirical comparison reports that paired test-example statistic directly rather than subtracting two marginal mean cardinalities. This isolates one-sided inflation caused by replacing the last-layer score with the terminal path maximum. For the six binary first-singleton cells, the returned-prefix set and $\C_L$ can differ only on an empty-terminal event; the measured empty-terminal rates in Appendix~\ref{app:cell-table} therefore make that separate prefix-versus-terminal gap at most a few thousandths in mean cardinality.
\subsection{Numerical scale of the efficiency bound}
\label{sec:theorem-scale}
The admissibility condition $m_t\leq n-k$ forces the empirical threshold into the upper-$\alpha$ tail of the early-over-terminal excess $D$. Evaluated at the smallest admissible $t_\star$, the held-out quantity $w_{t_\star}\sum_y\widehat M_y$ is $\ensuremath{0.016}$--$\ensuremath{0.086}$ across the eight encoder cells, compared with observed test means $\ensuremath{0.01}$--$\ensuremath{0.08}$ for $(|\C_L^{\rm path}|-|\C_L^{\rm last}|)_+$. The design-role slope estimates are held fixed when test-role cardinality summaries change; in particular, the four RTE/BoolQ values are not rescaled to track the observed gap. The cell-wise RHS/observed ratio is $\ensuremath{1.07}$--$\ensuremath{1.73}$. Appendix~\ref{app:theorem-scale-full} reports $t_\star$, the order-statistic gap, the held-out local-mass estimate, and all cell-wise values.
\subsection{Policy-uniform frontier and sparse schedules}
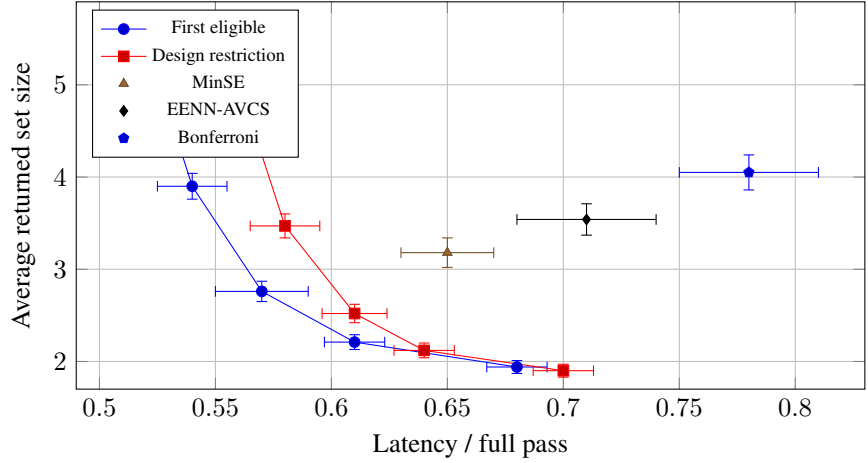
\begin{figure}[t]
\centering
\begin{tikzpicture}
\begin{axis}[
width=0.86\linewidth,
height=0.48\linewidth,
xlabel={Latency / full pass},
ylabel={Average returned set size},
xmin=0.49,xmax=0.83,
ymin=1.7,ymax=5.9,
grid=major,
legend style={font=\scriptsize,at={(0.02,0.98)},anchor=north west}
]
\addplot+[mark=*,error bars/.cd,x dir=both,x explicit,y dir=both,y explicit] coordinates {
(0.68,1.94) +- (0.013,0.07)
(0.61,2.21) +- (0.013,0.08)
(0.57,2.76) +- (0.020,0.11)
(0.54,3.90) +- (0.015,0.14)
(0.52,5.48) +- (0.017,0.18)
};
\addlegendentry{First eligible}
\addplot+[mark=square*,error bars/.cd,x dir=both,x explicit,y dir=both,y explicit] coordinates {
(0.70,1.90) +- (0.013,0.07)
(0.64,2.12) +- (0.013,0.08)
(0.61,2.52) +- (0.014,0.10)
(0.58,3.47) +- (0.015,0.13)
(0.56,5.01) +- (0.017,0.17)
};
\addlegendentry{Design restriction}
\addplot+[only marks,mark=triangle*,error bars/.cd,x dir=both,x explicit,y dir=both,y explicit] coordinates {
(0.65,3.18) +- (0.020,0.16)
};
\addlegendentry{MinSE}
\addplot+[only marks,mark=diamond*,error bars/.cd,x dir=both,x explicit,y dir=both,y explicit] coordinates {
(0.71,3.54) +- (0.030,0.17)
};
\addlegendentry{EENN-AVCS}
\addplot+[only marks,mark=pentagon*,error bars/.cd,x dir=both,x explicit,y dir=both,y explicit] coordinates {
(0.78,4.05) +- (0.030,0.19)
};
\addlegendentry{Bonferroni}
\end{axis}
\end{tikzpicture}
\caption{CIFAR-100 latency--set-size tradeoff for the explicit $(n,L)=(\ensuremath{500},\ensuremath{24})$ configuration used in the scaling grid. Bars are paired $95\%$ hierarchical-bootstrap intervals over head seeds and calibration subsamples. For each budget $b\in\ensuremath{\{1,2,4,8,16\}}$, first-eligible is the earliest admissible stop; the design-chosen slack policy is a restriction and therefore lies weakly to its right in latency while returning smaller sets. The MinSE, EENN-AVCS, and Bonferroni markers are exactly the $(500,24)$ entries of Appendix~\ref{app:scaling-grid}. Exact coverage, empty-terminal rate, and frontier values are reported in Appendix~\ref{app:policy-frontier}.}
\label{fig:frontier}
\end{figure}

The cardinality sweep is degenerate on binary tasks, so the main operating frontier uses CIFAR-100, where $K=100$. To keep every baseline finite and make the frontier directly reproducible from the scaling grid, we fix $(n,L)=(\ensuremath{500},\ensuremath{24})$. Figure~\ref{fig:frontier} then sweeps only the deployment set budget while keeping that terminal calibration fixed. First-eligible stopping is latency-minimal among policies that must wait until the current pathwise set has cardinality at most $b$: any policy obtained by adding exit-specific slack constraints is a restriction of first-eligible and therefore cannot stop earlier on any input. The design-chosen policy uses such constraints to trade additional latency for smaller returned sets. Because both policies return supersets of the same terminal set, every displayed operating point inherits the same conformal guarantee; the contribution of policy-uniformity is post-calibration flexibility, not a claim that a restriction can dominate first-eligible latency.
The frontier is evaluated at the same $(n,L)=(\ensuremath{500},\ensuremath{24})$ configuration as the corresponding scaling-grid row. At budget $b=\ensuremath{4}$, first-eligible stopping attains coverage $\ensuremath{0.902}$, average set size $\ensuremath{2.76}$, and latency fraction $\ensuremath{0.57}$. The design-chosen restriction returns a smaller set, $\ensuremath{2.52}$ on average, at coverage $\ensuremath{0.901}$ and latency $\ensuremath{0.61}$; the increased latency is required by the pointwise ordering of stopping times. At the same calibration size and retained-exit count, MinSE valid selection has set size/latency $\ensuremath{3.18/0.65}$, EENN-AVCS has $\ensuremath{3.54/0.71}$, and Bonferroni has $\ensuremath{4.05/0.78}$, exactly matching Appendix~\ref{app:scaling-grid}. The same fitted $\widehat q$ supports all five pathwise budgets and both pathwise policies. In contrast, the supervised risk-control baseline is policy-specific: each changed deployment budget is re-fit on its calibration role before evaluation.

The sparse encoder schedule is the hardest case for the no-$\alpha/L$ claim because it retains only three or four candidate exits. We therefore compare it directly with Bonferroni at the same schedules rather than only against the dense network. With $L=\ensuremath{3}$, pathwise coverage/set/latency are $\ensuremath{0.901/1.34/0.60}$ versus $\ensuremath{0.944/1.52/0.68}$ for Bonferroni; with $L=\ensuremath{4}$ they are $\ensuremath{0.902/1.36/0.62}$ versus $\ensuremath{0.949/1.60/0.72}$. Thus the multiplicity correction remains measurable even when the schedule is already sparse.
\subsection{Scaling in candidate exits and calibration size}
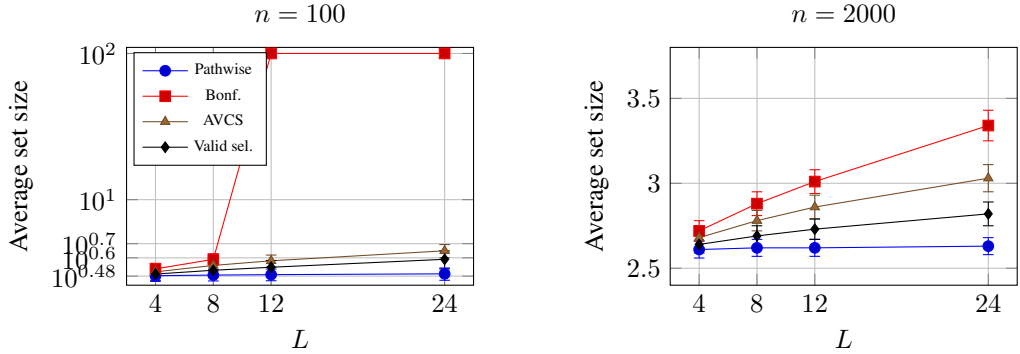
\begin{figure}[t]
\centering
\begin{minipage}{0.47\linewidth}
\centering
\begin{tikzpicture}
\begin{axis}[
width=0.94\linewidth,height=0.72\linewidth,
xlabel={$L$},ylabel={Average set size},
xtick={4,8,12,24},
ymode=log,
ymin=2.6,ymax=110,
ytick={3,4,5,10,100},
grid=major,title={$n=100$},
legend style={font=\tiny,at={(0.02,0.98)},anchor=north west}
]
\addplot+[mark=*,error bars/.cd,y dir=both,y explicit] coordinates {(4,3.02) +- (0,0.26)
(8,3.05) +- (0,0.27)
(12,3.07) +- (0,0.28)
(24,3.11) +- (0,0.30)};\addlegendentry{Pathwise}
\addplot+[mark=square*,error bars/.cd,y dir=both,y explicit] coordinates {(4,3.36) +- (0,0.31)
(8,3.91) +- (0,0.35)
(12,100) +- (0,0)
(24,100) +- (0,0)};\addlegendentry{Bonf.}
\addplot+[mark=triangle*,error bars/.cd,y dir=both,y explicit] coordinates {(4,3.21) +- (0,0.30)
(8,3.55) +- (0,0.33)
(12,3.82) +- (0,0.36)
(24,4.47) +- (0,0.46)};\addlegendentry{AVCS}
\addplot+[mark=diamond*,error bars/.cd,y dir=both,y explicit] coordinates {(4,3.10) +- (0,0.29)
(8,3.29) +- (0,0.34)
(12,3.45) +- (0,0.39)
(24,3.91) +- (0,0.55)};\addlegendentry{Valid sel.}
\end{axis}
\end{tikzpicture}
\end{minipage}\hfill
\begin{minipage}{0.47\linewidth}
\centering
\begin{tikzpicture}
\begin{axis}[
width=0.94\linewidth,height=0.72\linewidth,
xlabel={$L$},ylabel={Average set size},
xtick={4,8,12,24},
ymin=2.4,ymax=3.8,
grid=major,title={$n=2000$}
]
\addplot+[mark=*,error bars/.cd,y dir=both,y explicit] coordinates {(4,2.61) +- (0,0.05)
(8,2.62) +- (0,0.05)
(12,2.62) +- (0,0.05)
(24,2.63) +- (0,0.05)};
\addplot+[mark=square*,error bars/.cd,y dir=both,y explicit] coordinates {(4,2.72) +- (0,0.06)
(8,2.88) +- (0,0.07)
(12,3.01) +- (0,0.07)
(24,3.34) +- (0,0.09)};
\addplot+[mark=triangle*,error bars/.cd,y dir=both,y explicit] coordinates {(4,2.68) +- (0,0.06)
(8,2.78) +- (0,0.06)
(12,2.86) +- (0,0.07)
(24,3.03) +- (0,0.08)};
\addplot+[mark=diamond*,error bars/.cd,y dir=both,y explicit] coordinates {(4,2.64) +- (0,0.05)
(8,2.69) +- (0,0.06)
(12,2.73) +- (0,0.06)
(24,2.82) +- (0,0.07)};
\end{axis}
\end{tikzpicture}
\end{minipage}
\caption{Scaling with the number of candidate exits on CIFAR-100. Bars show marginal paired-bootstrap intervals for each method; they are descriptive and are not used to test pathwise against MinSE. The left panel uses $n=\ensuremath{100}$ and a log $y$-axis because Bonferroni has rank $n+1$ at $L=\ensuremath{12}$ and $L=\ensuremath{24}$, forcing the returned set to all $\ensuremath{100}$ classes. The right panel uses $n=\ensuremath{2000}$. Paired pathwise--MinSE difference intervals, which exploit the shared calibration subsample and test set, appear in Appendix~\ref{app:scaling-grid}.}
\label{fig:scaling}
\end{figure}

The layer-count effect is the empirical consequence most directly tied to the theory. Figure~\ref{fig:scaling} holds the frozen ViT-L/16 score path fixed, varies the number of retained candidate exits, and repeats the comparison at small and large calibration sizes. Pathwise calibration uses one terminal rank regardless of $L$; Bonferroni protects all exits simultaneously; EENN-AVCS runs the classification AVCS recursion over the retained exits; and the valid-selection line is the MinSE instantiation specified in Appendix~\ref{app:baseline-adaptation}. These are concrete comparator implementations rather than claims about every possible AVCS or stable selector. The point estimates separate most strongly when $n$ is small and $L$ is large, but uncertainty is also largest there.
At $n=\ensuremath{100}$, increasing $L$ from $\ensuremath{4}$ to $\ensuremath{24}$ changes pathwise average set size by only $\ensuremath{0.09}$. Bonferroni is finite at $L=\ensuremath{4}$ and $L=\ensuremath{8}$ but becomes exactly degenerate at $L=\ensuremath{12}$ and $L=\ensuremath{24}$ because its conformal rank is $n+1$; every set is then the full $\ensuremath{100}$-class label space and latency is $\ensuremath{1.00}$ of a full pass. EENN-AVCS and MinSE increase by $\ensuremath{1.26}$ and $\ensuremath{0.81}$ over the same $L=\ensuremath{4}$--$\ensuremath{24}$ range. Inference against MinSE uses paired differences, not overlap of marginal intervals: at $(n,L)=(\ensuremath{100},\ensuremath{24})$, MinSE minus pathwise set size is $\ensuremath{0.80}$ with paired $95\%$ interval $\ensuremath{[0.47,1.13]}$. The paired grid shows the same long-path separation at small calibration sizes once $L$ is sufficiently large. At $n=\ensuremath{2000}$ and $L=\ensuremath{24}$, the paired difference is $\ensuremath{0.19\ [0.12,0.26]}$. Table~\ref{tab:vision-replication} shows that the qualitative ordering persists when depth is changed by training separate paths rather than retaining exits from one frozen path.

Calibration size is a separate source of conservatism. The marginal guarantee remains exact, but a
small calibration set produces a coarse order statistic, a weaker training-conditional floor, and
greater variation across calibration draws. The headline RTE result now uses $\ensuremath{498}$ calibration examples from the official training split, so the calibration-size stress test is reported separately rather than embedded in the benchmark split. Table~\ref{tab:calibration-scaling} therefore
reports both empirical coverage variation and the lower bounds that are valid for the realized
calibration procedure.

At $n=100$, coverage SD is $0.028$ and the worst draw is $0.841$, with Beta
and DKW floors $0.848$ and $0.778$.  At $n=1500$, the SD falls to $0.006$ and
the worst draw rises to $0.888$, while the floors reach $0.887$ and $0.868$.

Thus mean coverage near $0.90$ coexists with unstable realized calibration at
small $n$, even though marginal validity remains exact.

Together, the matched and ablation results show that pathwise calibration can
meet its target and exit early on small-label encoder tasks.  The decoder study
below identifies the complementary regime in which exact coverage forces
near-full-depth execution.

The score and schedule analysis covers $-\log p$, design-split
transformations, APS, RAPS, first-eligible exit, and sparse schedules
\citep{romano2020classification,angelopoulos2021uncertainty}.

Appendix~\ref{app:decoder-boundary} reports the full-vocabulary boundary.
\subsection{Uncertainty and paired comparisons}
All method comparisons are paired at the level at which the data roles actually vary. SST-2, MNLI, and BoolQ resample training seeds and paired calibration--test roles; RTE keeps its calibration/test partition fixed and resamples head-training seeds and examples from the fixed development test set. The complete decision rules, familywise procedure, and latency-comparison protocol are given in Appendix~\ref{app:uncertainty}.
\section{Discussion, Conclusion, and Limitations}
\label{sec:discussion}\label{sec:conclusion}\label{sec:limitations}
Pathwise calibration makes adaptive-depth selection valid through one terminal conformal rank: every admissible returned set contains the calibrated terminal set. Its practical advantage is clearest when many exits share limited calibration data or deployment budgets change after calibration; on short encoder paths, performance remains close to MinSE valid selection. First-eligible stopping is latency-minimal for a fixed cardinality budget, while additional restrictions trade latency for smaller sets. The efficiency theorem explains the cost of the path maximum through early-over-terminal score excess and local threshold mass, although empirical slope estimates do not certify the population constants. Coverage remains marginal under exchangeability and frozen scores, and a weak shallow head can make the sets uninformative. Mondrian protection additionally requires stable within-group score laws and adequate calibration counts. The retained-exit sweep isolates candidate-count effects; separately trained vision replications broaden the evidence but confound depth with capacity. Full-vocabulary decoder sets remain operationally inert and protect next-token coverage rather than sequence correctness. These boundaries favor deployment where shallow heads produce actionable sets, with the full interpretation and limitations retained in Appendix~\ref{app:closing-analysis}.
\section*{Reproducibility Statement}
Appendix~\ref{app:algorithm} specifies execution and fallback behavior. Appendix~\ref{app:proofs} gives the marginal, conditional, Mondrian, and hybrid arguments, and Appendix~\ref{app:efficiency-proof} proves the path-maximum efficiency bound. Appendix~\ref{app:protocol} fixes data roles, paired resampling, familywise correction, latency scope, and method-specific targets. Appendices~\ref{app:cell-table}--\ref{app:scaling-grid} give the exact values behind the per-cell, policy-frontier, and $L$--$n$ figures. The timing record includes device, precision, batch size, input size, warm-up, synchronization, and repetition count. The 2025--2026 bibliography entries were cross-checked against venue, proceedings, or arXiv records.
\clearpage
\section*{AI Use Statement}
Generative AI assisted with manuscript organization, prose revision, and mathematical review. The authors remain responsible for independently verifying all claims, citations, proofs, protocols, and numerical results before submission.
\bibliographystyle{iclr2027_conference}
\bibliography{references}

@book{vovk2005algorithmic,
  title={Algorithmic Learning in a Random World},
  author={Vovk, Vladimir and Gammerman, Alex and Shafer, Glenn},
  year={2005},
  publisher={Springer}
}

@article{shafer2008tutorial,
  title={A Tutorial on Conformal Prediction},
  author={Shafer, Glenn and Vovk, Vladimir},
  journal={Journal of Machine Learning Research},
  volume={9},
  pages={371--421},
  year={2008}
}

@inproceedings{papadopoulos2002inductive,
  title={Inductive Confidence Machines for Regression},
  author={Papadopoulos, Harris and Proedrou, Kostas and Vovk, Vladimir and Gammerman, Alex},
  booktitle={European Conference on Machine Learning},
  pages={345--356},
  year={2002},
  publisher={Springer}
}

@article{lei2018distribution,
  title={Distribution-Free Predictive Inference for Regression},
  author={Lei, Jing and G'Sell, Max and Rinaldo, Alessandro and Tibshirani, Ryan J. and Wasserman, Larry},
  journal={Journal of the American Statistical Association},
  volume={113},
  number={523},
  pages={1094--1111},
  year={2018}
}

@article{vovk2015crossconformal,
  title={Cross-Conformal Predictors},
  author={Vovk, Vladimir},
  journal={Annals of Mathematics and Artificial Intelligence},
  volume={74},
  pages={9--28},
  year={2015}
}

@article{barber2021predictive,
  title={Predictive Inference with the Jackknife+},
  author={Barber, Rina Foygel and Cand{\`e}s, Emmanuel J. and Ramdas, Aaditya and Tibshirani, Ryan J.},
  journal={The Annals of Statistics},
  volume={49},
  number={1},
  pages={486--507},
  year={2021}
}

@inproceedings{romano2020classification,
  title={Classification with Valid and Adaptive Coverage},
  author={Romano, Yaniv and Sesia, Matteo and Cand{\`e}s, Emmanuel J.},
  booktitle={Advances in Neural Information Processing Systems},
  volume={33},
  pages={3581--3591},
  year={2020}
}

@inproceedings{angelopoulos2021uncertainty,
  title={Uncertainty Sets for Image Classifiers Using Conformal Prediction},
  author={Angelopoulos, Anastasios N. and Bates, Stephen and Jordan, Michael I. and Malik, Jitendra},
  booktitle={International Conference on Learning Representations},
  year={2021}
}

@article{sadinle2019least,
  title={Least Ambiguous Set-Valued Classifiers with Bounded Error Levels},
  author={Sadinle, Mauricio and Lei, Jing and Wasserman, Larry},
  journal={Journal of the American Statistical Association},
  volume={114},
  number={525},
  pages={223--234},
  year={2019}
}

@article{angelopoulos2023gentle,
  title={Conformal Prediction: A Gentle Introduction},
  author={Angelopoulos, Anastasios N. and Bates, Stephen},
  journal={Foundations and Trends in Machine Learning},
  volume={16},
  number={4},
  pages={494--591},
  year={2023}
}

@article{gupta2022nested,
  title={Nested Conformal Prediction and Quantile Out-of-Bag Ensemble Methods},
  author={Gupta, Chirag and Kuchibhotla, Arun K. and Ramdas, Aaditya},
  journal={Pattern Recognition},
  volume={127},
  pages={108496},
  year={2022}
}

@inproceedings{angelopoulos2024conformalrisk,
  title={Conformal Risk Control},
  author={Angelopoulos, Anastasios N. and Bates, Stephen and Fisch, Adam and Lei, Lihua and Schuster, Tal},
  booktitle={International Conference on Learning Representations},
  year={2024}
}

@article{bates2021distribution,
  title={Distribution-Free, Risk-Controlling Prediction Sets},
  author={Bates, Stephen and Angelopoulos, Anastasios N. and Lei, Lihua and Malik, Jitendra and Jordan, Michael I.},
  journal={Journal of the ACM},
  volume={68},
  number={6},
  pages={1--34},
  year={2021}
}

@article{angelopoulos2021learn,
  title={Learn Then Test: Calibrating Predictive Algorithms to Achieve Risk Control},
  author={Angelopoulos, Anastasios N. and Bates, Stephen and Cand{\`e}s, Emmanuel J. and Jordan, Michael I. and Lei, Lihua},
  journal={The Annals of Applied Statistics},
  volume={19},
  number={2},
  year={2025},
  note={First circulated as arXiv:2110.01052}
}

@article{shihab2026continual,
  title={Continual Calibration: Coverage Can Collapse Before Accuracy in Lifelong LLM Fine-Tuning},
  author={Shihab, Ibne Farabi and Akter, Sanjeda and Sharma, Anuj},
  journal={arXiv preprint arXiv:2604.23987},
  year={2026}
}

@article{vovk2012conditional,
  title={Conditional Validity of Inductive Conformal Predictors},
  author={Vovk, Vladimir},
  journal={Proceedings of Machine Learning Research},
  volume={25},
  pages={475--490},
  year={2012}
}

@techreport{vovk2003mondrian,
  title={Mondrian Confidence Machine},
  author={Vovk, Vladimir and Lindsay, David and Nouretdinov, Ilia and Gammerman, Alex},
  institution={Royal Holloway, University of London},
  number={On-line Compression Modelling Working Paper 4},
  year={2003}
}

@inproceedings{tibshirani2019conformal,
  title={Conformal Prediction Under Covariate Shift},
  author={Tibshirani, Ryan J. and Barber, Rina Foygel and Cand{\`e}s, Emmanuel J. and Ramdas, Aaditya},
  booktitle={Advances in Neural Information Processing Systems},
  volume={32},
  year={2019}
}

@article{barber2023conformal,
  title={Conformal Prediction Beyond Exchangeability},
  author={Barber, Rina Foygel and Cand{\`e}s, Emmanuel J. and Ramdas, Aaditya and Tibshirani, Ryan J.},
  journal={The Annals of Statistics},
  volume={51},
  number={2},
  pages={816--845},
  year={2023}
}

@article{yang2025selection,
  title={Selection and Aggregation of Conformal Prediction Sets},
  author={Yang, Yachong and Kuchibhotla, Arun Kumar},
  journal={Journal of the American Statistical Association},
  volume={120},
  number={549},
  pages={435--447},
  year={2025}
}

@article{sarkar2023postselection,
  title={Post-Selection Inference for Conformal Prediction: Trading Off Coverage for Precision},
  author={Sarkar, Siddhaarth and Kuchibhotla, Arun Kumar},
  journal={arXiv preprint arXiv:2304.06158},
  year={2023}
}

@inproceedings{hegazy2025valid,
  title={Valid Selection among Conformal Sets},
  author={Hegazy, Mahmoud and Aolaritei, Liviu and Jordan, Michael I. and Dieuleveut, Aymeric},
  booktitle={Advances in Neural Information Processing Systems},
  volume={38},
  year={2025}
}

@article{bengio2015conditional,
  title={Conditional Computation in Neural Networks for Faster Models},
  author={Bengio, Emmanuel and Bacon, Pierre-Luc and Pineau, Joelle and Precup, Doina},
  journal={arXiv preprint arXiv:1511.06297},
  year={2015}
}

@article{graves2016adaptive,
  title={Adaptive Computation Time for Recurrent Neural Networks},
  author={Graves, Alex},
  journal={arXiv preprint arXiv:1603.08983},
  year={2016}
}

@inproceedings{figurnov2017spatially,
  title={Spatially Adaptive Computation Time for Residual Networks},
  author={Figurnov, Michael and Collins, Maxwell D. and Zhu, Yukun and Zhang, Li and Huang, Jonathan and Vetrov, Dmitry and Salakhutdinov, Ruslan},
  booktitle={Advances in Neural Information Processing Systems},
  volume={30},
  year={2017}
}

@inproceedings{teerapittayanon2016branchynet,
  title={BranchyNet: Fast Inference via Early Exiting from Deep Neural Networks},
  author={Teerapittayanon, Surat and McDanel, Bradley and Kung, H. T.},
  booktitle={International Conference on Pattern Recognition},
  pages={2464--2469},
  year={2016}
}

@article{han2022survey,
  title={Dynamic Neural Networks: A Survey},
  author={Han, Yizeng and Huang, Gao and Song, Shiji and Yang, Le and Wang, Honghui and Wang, Yulin},
  journal={IEEE Transactions on Pattern Analysis and Machine Intelligence},
  volume={44},
  number={11},
  pages={7436--7456},
  year={2022}
}

@article{scardapane2020why,
  title={Why Should We Add Early Exits to Neural Networks?},
  author={Scardapane, Simone and Scarpiniti, Michele and Baccarelli, Enzo and Uncini, Aurelio},
  journal={Cognitive Computation},
  volume={12},
  pages={954--966},
  year={2020}
}

@inproceedings{xin2020deebert,
  title={DeeBERT: Dynamic Early Exiting for Accelerating BERT Inference},
  author={Xin, Ji and Tang, Raphael and Lee, Jaejun and Yu, Yaoliang and Lin, Jimmy},
  booktitle={Annual Meeting of the Association for Computational Linguistics},
  pages={2246--2251},
  year={2020}
}

@inproceedings{liu2020fastbert,
  title={FastBERT: A Self-Distilling BERT with Adaptive Inference Time},
  author={Liu, Weijie and Zhou, Peng and Zhao, Zhe and Wang, Zhiruo and Deng, Haotang and Ju, Qi},
  booktitle={Annual Meeting of the Association for Computational Linguistics},
  pages={6035--6044},
  year={2020}
}

@inproceedings{zhou2020bert,
  title={BERT Loses Patience: Fast and Robust Inference with Early Exit},
  author={Zhou, Wangchunshu and Xu, Canwen and Ge, Tao and McAuley, Julian and Xu, Ke and Wei, Furu},
  booktitle={Advances in Neural Information Processing Systems},
  volume={33},
  pages={18330--18341},
  year={2020}
}

@inproceedings{schwartz2020right,
  title={The Right Tool for the Job: Matching Model and Instance Complexities},
  author={Schwartz, Roy and Stanovsky, Gabriel and Swayamdipta, Swabha and Dodge, Jesse and Smith, Noah A.},
  booktitle={Annual Meeting of the Association for Computational Linguistics},
  pages={6640--6651},
  year={2020}
}

@inproceedings{elbayad2020depth,
  title={Depth-Adaptive Transformer},
  author={Elbayad, Maha and Gu, Jiatao and Grave, Edouard and Auli, Michael},
  booktitle={International Conference on Learning Representations},
  year={2020}
}

@inproceedings{huang2018msdnet,
  title={Multi-Scale Dense Networks for Resource Efficient Image Classification},
  author={Huang, Gao and Chen, Danlu and Li, Tianhong and Wu, Felix and van der Maaten, Laurens and Weinberger, Kilian Q.},
  booktitle={International Conference on Learning Representations},
  year={2018}
}

@inproceedings{schuster2021consistent,
  title={Consistent Accelerated Inference via Confident Adaptive Transformers},
  author={Schuster, Tal and Fisch, Adam and Jaakkola, Tommi S. and Barzilay, Regina},
  booktitle={Conference on Empirical Methods in Natural Language Processing},
  pages={4962--4979},
  year={2021}
}

@inproceedings{schuster2022confident,
  title={Confident Adaptive Language Modeling},
  author={Schuster, Tal and Fisch, Adam and Gupta, Jai and Dehghani, Mostafa and Bahri, Dara and Tran, Vinh Q. and Tay, Yi and Metzler, Donald},
  booktitle={Advances in Neural Information Processing Systems},
  volume={35},
  pages={17456--17472},
  year={2022}
}

@inproceedings{jazbec2024fast,
  title={Fast yet Safe: Early-Exiting with Risk Control},
  author={Jazbec, Metod and Timans, Alexander and Had{\v z}i Veljkovi{\'c}, Tin and Sakmann, Kaspar and Zhang, Dan and Naesseth, Christian A. and Nalisnick, Eric},
  booktitle={Advances in Neural Information Processing Systems},
  volume={37},
  year={2024}
}

@inproceedings{jazbec2024nested,
  title={Early-Exit Neural Networks with Nested Prediction Sets},
  author={Jazbec, Metod and Forr{\'e}, Patrick and Mandt, Stephan and Zhang, Dan and Nalisnick, Eric},
  booktitle={Proceedings of the Fortieth Conference on Uncertainty in Artificial Intelligence},
  series={Proceedings of Machine Learning Research},
  volume={244},
  pages={1780--1796},
  year={2024}
}

@article{kubaty2026rethinking,
  title={Rethinking Calibration for Early-Exit Neural Networks},
  author={Kubaty, Piotr and Szatkowski, Filip and Choczy{\'n}ski, Grzegorz and Nalisnick, Eric and W{\'o}jcik, Bartosz},
  journal={arXiv preprint arXiv:2508.21495},
  year={2026}
}

@article{guo2026earlyexit,
  title={When Do Early-Exit Networks Generalize? A PAC-Bayesian Theory of Adaptive Depth},
  author={Guo, Dongxin and Wu, Jikun and Yiu, Siu Ming},
  journal={arXiv preprint arXiv:2604.15764},
  year={2026}
}

@inproceedings{geifman2017selective,
  title={Selective Classification for Deep Neural Networks},
  author={Geifman, Yonatan and El-Yaniv, Ran},
  booktitle={Advances in Neural Information Processing Systems},
  volume={30},
  year={2017}
}

@inproceedings{geifman2019selectivenet,
  title={SelectiveNet: A Deep Neural Network with an Integrated Reject Option},
  author={Geifman, Yonatan and El-Yaniv, Ran},
  booktitle={International Conference on Machine Learning},
  series={Proceedings of Machine Learning Research},
  volume={97},
  pages={2151--2159},
  year={2019}
}

@article{elyaniv2010foundations,
  title={On the Foundations of Noise-Free Selective Classification},
  author={El-Yaniv, Ran and Wiener, Yair},
  journal={Journal of Machine Learning Research},
  volume={11},
  pages={1605--1641},
  year={2010}
}

@article{guo2017calibration,
  title={On Calibration of Modern Neural Networks},
  author={Guo, Chuan and Pleiss, Geoff and Sun, Yu and Weinberger, Kilian Q.},
  journal={Proceedings of Machine Learning Research},
  volume={70},
  pages={1321--1330},
  year={2017}
}

@article{maltoudoglou2020conformal,
  title={Well-Calibrated Confidence Measures for Multi-Label Text Classification with a Large Number of Labels},
  author={Maltoudoglou, Lysimachos and Paisios, Andreas and Lenc, Ladislav and Mart{\'i}nek, Ji{\v r}{\'i} and Kr{\'a}l, Pavel and Papadopoulos, Harris},
  journal={Pattern Recognition},
  volume={122},
  pages={108271},
  year={2022}
}

@article{kumar2023conformal,
  title={Conformal Prediction with Large Language Models for Multi-Choice Question Answering},
  author={Kumar, Bhawesh and Lu, Charlie and Gupta, Gauri and Palepu, Anil and Bellamy, David and Raskar, Ramesh and Beam, Andrew},
  journal={arXiv preprint arXiv:2305.18404},
  year={2023}
}

@inproceedings{quach2024conformal,
  title={Conformal Language Modeling},
  author={Quach, Victor and Fisch, Adam and Schuster, Tal and Yala, Adam and Sohn, Jae Ho and Jaakkola, Tommi S. and Barzilay, Regina},
  booktitle={International Conference on Learning Representations},
  year={2024}
}

@article{campos2024conformalnlp,
  title={Conformal Prediction for Natural Language Processing: A Survey},
  author={Campos, Margarida and Farinhas, Ant{\'o}nio and Zerva, Chrysoula and Figueiredo, M{\'a}rio A. T. and Martins, Andr{\'e} F. T.},
  journal={Transactions of the Association for Computational Linguistics},
  volume={12},
  pages={1497--1516},
  year={2024}
}

@inproceedings{fisch2021efficient,
  title={Efficient Conformal Prediction via Cascaded Inference with Expanded Admission},
  author={Fisch, Adam and Schuster, Tal and Jaakkola, Tommi S. and Barzilay, Regina},
  booktitle={International Conference on Learning Representations},
  year={2021}
}

@inproceedings{wang2018glue,
  title={GLUE: A Multi-Task Benchmark and Analysis Platform for Natural Language Understanding},
  author={Wang, Alex and Singh, Amanpreet and Michael, Julian and Hill, Felix and Levy, Omer and Bowman, Samuel R.},
  booktitle={International Conference on Learning Representations},
  year={2019}
}

@inproceedings{wang2019superglue,
  title={SuperGLUE: A Stickier Benchmark for General-Purpose Language Understanding Systems},
  author={Wang, Alex and Pruksachatkun, Yada and Nangia, Nikita and Singh, Amanpreet and Michael, Julian and Hill, Felix and Levy, Omer and Bowman, Samuel R.},
  booktitle={Advances in Neural Information Processing Systems},
  volume={32},
  year={2019}
}

@inproceedings{vaswani2017attention,
  title={Attention Is All You Need},
  author={Vaswani, Ashish and Shazeer, Noam and Parmar, Niki and Uszkoreit, Jakob and Jones, Llion and Gomez, Aidan N. and Kaiser, {\L}ukasz and Polosukhin, Illia},
  booktitle={Advances in Neural Information Processing Systems},
  volume={30},
  year={2017}
}

@article{sanh2019distilbert,
  title={DistilBERT, a Distilled Version of BERT: Smaller, Faster, Cheaper and Lighter},
  author={Sanh, Victor and Debut, Lysandre and Chaumond, Julien and Wolf, Thomas},
  journal={arXiv preprint arXiv:1910.01108},
  year={2019}
}

@inproceedings{devlin2019bert,
  title={BERT: Pre-Training of Deep Bidirectional Transformers for Language Understanding},
  author={Devlin, Jacob and Chang, Ming-Wei and Lee, Kenton and Toutanova, Kristina},
  booktitle={Conference of the North American Chapter of the Association for Computational Linguistics},
  pages={4171--4186},
  year={2019}
}

@article{dvoretzky1956asymptotic,
  title={Asymptotic Minimax Character of the Sample Distribution Function and of the Classical Multinomial Estimator},
  author={Dvoretzky, Aryeh and Kiefer, Jack and Wolfowitz, Jacob},
  journal={The Annals of Mathematical Statistics},
  volume={27},
  number={3},
  pages={642--669},
  year={1956}
}
\clearpage
\appendix
\section{Adaptive selection can multiply miscoverage}
\label{app:selection}
Let $Y=1$ almost surely, and let $U\sim\mathrm{Unif}[0,1]$ be part of $X$. Choose disjoint intervals
$I_1,\ldots,I_L$ of length $\alpha$, with $L\alpha\leq1$. At exit $\ell$, return
$\C_\ell(U)=\{0\}$ if $U\in I_\ell$ and $\C_\ell(U)=\{0,1\}$ otherwise. The set at any fixed exit
misses the true label only on $I_\ell$, so its coverage is $1-\alpha$.

An earliest-singleton rule returns $\{0\}$ whenever $U\in\cup_{\ell=1}^L I_\ell$. The intervals
are disjoint, and therefore
\[
 \Pr\{Y\notin\C_{\widehat\ell(U)}(U)\}
 =\Pr\left\{U\in\bigcup_{\ell=1}^L I_\ell\right\}
 =L\alpha.
\]
The rule uses only $U$ and the set sizes; it never observes the unknown label. Fixed-layer marginal
validity alone therefore cannot justify validity after adaptive selection.
\section{Selection-safe algorithm}
\label{app:algorithm}
\begin{enumerate}
\item Freeze the backbone, exit heads, exit schedule, candidatewise scores $r_1,\ldots,r_L$, and
any transformation learned on $\D_{\mathrm{des}}$.
\item For every calibration pair, compute
$S_i=\max_{\ell\leq L}r_\ell(X_i,Y_i)$.
\item Set $k=\lceil(n+1)(1-\alpha)\rceil$ and let $\widehat q$ be the $k$th calibration order
statistic, using $+\infty$ when $k=n+1$.
\item For a test input $x$, update
$A_\ell(x,y)=\max\{A_{\ell-1}(x,y),r_\ell(x,y)\}$ for every candidate $y$, with
$A_0(x,y)=-\infty$.
\item Form $\C_\ell(x)=\{y:A_\ell(x,y)\leq\widehat q\}$ and return the first set satisfying
$1\leq|\C_\ell(x)|\leq b$.
\item If no exit fires, return $\C_L(x)$ when it is nonempty and $\Y$ otherwise. Returning $\Y$
only enlarges the terminal set.
\end{enumerate}
The cardinality rule in Step 5 is optional. Any label-free rule is valid when the returned set
contains $\C_L(x)$.
\section{Proofs}
\label{app:proofs}
\subsection{Proof of Theorem~\ref{thm:coverage}}
Conditional on every object fitted before final calibration, the terminal scores
\[
 S_i=A_L(X_i,Y_i),\quad i=1,\ldots,n,
 \qquad S_{n+1}=A_L(X,Y)
\]
are exchangeable. With $k=\lceil(n+1)(1-\alpha)\rceil$ and the standard $+\infty$ convention, the
split-conformal rank argument gives
\[
 \Pr\{S_{n+1}\leq\widehat q\}\geq\frac{k}{n+1}\geq1-\alpha.
\]
This event is exactly $\{Y\in\C_L(X)\}$. Since
$A_\ell(X,y)\leq A_L(X,y)$ for every $\ell$ and $y$, we have
$\C_\ell(X)\supseteq\C_L(X)$. The algorithm returns a prefix set or $\Y$, so
\[
 \{Y\in\C_L(X)\}
 \subseteq
 \{Y\in\C_{\widehat\ell(X)}(X)\}.
\]
The claimed bound follows without independence across exits.
\subsection{Training-conditional coverage}
Let $F$ be the conditional CDF of $A_L(X,Y)$. If $F$ is continuous, then
$F(S_1),\ldots,F(S_n)$ are independent uniforms and
\[
 F(\widehat q)\sim\mathrm{Beta}(k,n+1-k).
\]
The selected prefix has conditional coverage at least $F(\widehat q)$. Thus, with probability at
least $1-\delta$ over $\D_{\mathrm{cal}}$, its coverage is at least
$F^{-1}_{\mathrm{Beta}(k,n+1-k)}(\delta)$. Atoms and conservative tie handling make this lower bound
conservative.

For Equation~\eqref{eq:dkw}, the one-sided DKW inequality
\citep{dvoretzky1956asymptotic} gives
\[
 \Pr\left\{\sup_t(F_n(t)-F(t))>\epsilon\right\}\leq e^{-2n\epsilon^2}.
\]
At $\widehat q$, $F_n(\widehat q)\geq k/n\geq1-\alpha$. Taking
$\epsilon=\sqrt{\log(1/\delta)/(2n)}$ proves the stated floor.
\subsection{Mondrian coverage under group-prior shift}
For group $g$, calibrate $\widehat q_g$ from terminal scores whose calibration pair belongs to
$g$, and define
\[
 \C_\ell^{\mathrm M}(x)=
 \{y:A_\ell(x,y)\leq\widehat q_{G(x,y)}\}.
\]
Within-group exchangeability gives
$\Pr\{Y\in\C_L^{\mathrm M}(X)\mid G(X,Y)=g\}\geq1-\alpha$. Candidatewise nesting makes the
selected set a superset of the terminal set. If deployment weights are $\pi'(g)$ while the
conditional score laws remain fixed, then
\[
 \Pr_{\pi'}\{Y\in\C_{\widehat\ell(X)}^{\mathrm M}(X)\}
 \geq\sum_g\pi'(g)(1-\alpha)=1-\alpha.
\]
\subsection{Hybrid error budget}
The hybrid misses only if an early singleton is wrong or the fallback set misses:
\[
 \{Y\notin H(X)\}
 \subseteq
 \{E(X),h(X)\neq Y\}
 \cup
 \{Y\notin\C_{\widehat\ell(X)}(X)\}.
\]
The union bound yields error at most $\alpha_{\mathrm{pt}}+\alpha_{\mathrm{set}}$. If the component
bounds hold with calibration-draw probabilities $1-\delta_{\mathrm{pt}}$ and
$1-\delta_{\mathrm{set}}$, both hold with probability at least
$1-\delta_{\mathrm{pt}}-\delta_{\mathrm{set}}$.
\section{Additional guarantees, diagnostics, and boundary evaluations}
\label{app:additional-analysis}
\subsection{Conditional, group, and hybrid guarantees}\label{app:extensions}
The path construction inherits ordinary split-conformal training-conditional behavior, quantified in Proposition~\ref{prop:conditional}. If terminal
scores are continuous, the conditional coverage of $\C_L$ follows
$\mathrm{Beta}(k,n+1-k)$ when $k\leq n$; if $k=n+1$, the $+\infty$ convention gives coverage one.
\begin{proposition}[Training-conditional lower bounds]
\label{prop:conditional}
Let $p(\D_{\mathrm{cal}})$ denote the conditional coverage of the selected set. Under the conditions
of Theorem~\ref{thm:coverage}, with continuous terminal scores,
\[
 \Pr_{\D_{\mathrm{cal}}}\!\left\{
 p(\D_{\mathrm{cal}})\geq
 F^{-1}_{\mathrm{Beta}(k,n+1-k)}(\delta)
 \right\}\geq1-\delta.
\]
With ties, conservative inclusion preserves the lower bound. A simpler distribution-free statement
is
\begin{equation}
 \Pr_{\D_{\mathrm{cal}}}\!\left\{
 p(\D_{\mathrm{cal}})\geq
 1-\alpha-\sqrt{\frac{\log(1/\delta)}{2n}}
 \right\}\geq1-\delta.
 \label{eq:dkw}
\end{equation}
\end{proposition}
Only the terminal path score is calibrated, so the bound contains no $\log L$ term.

For a fixed group map $G(x,y)$, calibrating a separate terminal quantile $\widehat q_g$ at rank
$\lceil(n_g+1)(1-\alpha)\rceil$ within each group yields group-conditional coverage, with the same
$+\infty$ convention when the rank exceeds $n_g$. Label groups remain implementable: each candidate
$y$ is tested using $\widehat q_{G(x,y)}$, without observing the unknown test label. The guarantee
survives changes in group proportions when the within-group terminal-score laws remain unchanged. It
does not cover arbitrary covariate shift.
\begin{proposition}[Group-prior shift]
\label{prop:mondrian}
If calibration and test terminal scores are exchangeable within every fixed group, then
$\Pr\{Y\in\C^{\mathrm M}_{\widehat\ell(X)}(X)\mid G(X,Y)=g\}\geq1-\alpha$ for each group with a
valid quantile. Consequently, marginal coverage remains at least $1-\alpha$ under any change in the
group mixture that preserves the conditional terminal-score laws.
\end{proposition}
Finally, let $E(X)$ be an early point-commit event with prediction $h(X)$. If
$\Pr\{E(X),h(X)\neq Y\}\leq\alpha_{\mathrm{pt}}$ and the fallback path is calibrated at
$\alpha_{\mathrm{set}}$, define $H(X)=\{h(X)\}$ on $E(X)$ and use the selected path set otherwise.
\begin{proposition}[Budgeted hybrid]
\label{prop:hybrid}
The hybrid satisfies
$\Pr\{Y\notin H(X)\}\leq\alpha_{\mathrm{pt}}+\alpha_{\mathrm{set}}$.
High-probability component statements combine with the corresponding sum of calibration failure
probabilities.
\end{proposition}
Proposition~\ref{prop:hybrid} implies that a $0.1$ total error target requires an explicit split between the two branches. Table~\ref{tab:guarantees} summarizes the distinct guarantees, and Appendix~\ref{app:proofs} proves all three extensions.

Efficiency choices are made before final calibration. In addition to exit spacing and increasing score transformations, the design split may optimize a family of label-free policies. We use policies parameterized by a set budget $b$, an exit-dependent threshold slack $\gamma_\ell$, and an optional continuation probability $\rho_\ell$. A test input may stop at exit $\ell$ only when $1\leq|\C_\ell(x)|\leq b$ and the surviving-set slack
\[
s_\ell(x)=\widehat q-\max_{y\in\C_\ell(x)}A_\ell(x,y)
\]
exceeds $\gamma_\ell$; a randomized policy may then continue with probability $\rho_\ell$. The design split minimizes a provisional cost such as $\widehat{\mathbb E}[c_{\widehat\ell(X)}+\lambda|\C_{\widehat\ell(X)}(X)|]$. The selected score map and exit schedule are frozen before final calibration. After $\widehat q$ is fixed, however, $b$, $\gamma_\ell$, $\rho_\ell$, and load-dependent combinations of these quantities may change without recalibration because they select only among already frozen supersets of $\C_L$.
\subsection{Controlled theorem checks}
The controlled simulations isolate the two finite-sample extensions that are not visible from aggregate encoder coverage. The DKW experiment draws independent calibration sets at $n\in\{200,500,1500\}$ and $\delta\in\{0.1,0.05,0.01\}$, then evaluates the conditional coverage induced by the terminal path quantile. The group-prior experiment changes only the mixture weights of three fixed groups while preserving their conditional terminal-score laws. These experiments therefore test Proposition~\ref{prop:conditional} and Proposition~\ref{prop:mondrian} under their stated assumptions rather than serving as substitutes for the encoder benchmarks.

The nine-cell PAC results are reported in Appendix~\ref{app:statistical}. The pathwise floor contains no layer-count penalty, whereas the simultaneous twelve-exit comparison is uniformly lower. Under the theorem-aligned group-prior shift, per-group calibration raises the hard-group coverage from $0.6887$ to $0.9149$ and the shifted marginal coverage from $0.8112$ to $0.9129$.
\subsection{Decision rules and uncertainty}\label{app:uncertainty}
The evaluation separates validity, selection failure, and efficiency. For task--backbone cell $c$,
the paired selection gap is
\begin{equation}
 \Delta_c=\widehat{\Pr}\{Y\notin\C_{\widehat\ell}^{\mathrm{naive}}(X)\}
 -\widehat{\Pr}\{Y\notin\C_{\widehat\ell}^{\mathrm{path}}(X)\}.
 \label{eq:selection-gap}
\end{equation}
Both terms use the same trained heads, calibration draw, test examples, and stopping budget. We
report $\Delta_c$ with a paired interval rather than infer selection failure from two overlapping
marginal intervals. A hierarchical bootstrap first resamples training seeds. Its second stage resamples paired role splits for SST-2, MNLI, and BoolQ, but resamples fixed-test examples for RTE because RTE has only one role partition. The eight per-cell gap statistics form one declared family and use a hierarchical bootstrap maximum-absolute-studentized correction for $95\%$ familywise intervals; the aggregate cannot replace the per-cell results.

Coverage and efficiency answer different questions. Coverage is reported with its finite-sample
conformal target and the distribution across calibration draws. Latency and normalized FLOPs are
compared only among methods that meet their native controlled target, using the latency--set-size
frontier rather than one favorable operating point. A baseline whose full model cannot satisfy the
declared supervised-risk target is marked infeasible or abstaining, not counted as an undercoverage
win for the proposed method. Every latency interval uses paired executions with identical inputs,
hardware, precision, batch size, warm-up, and synchronization.
\subsection{Boundary tests}\label{app:boundary-details}
The group-shift experiment changes only group proportions after defining groups on observed
attributes or a design split. It reports per-group calibration counts, worst-group coverage, shifted
marginal coverage, set size, and latency. A shift that changes the within-group score law is labeled
as robustness rather than theorem-aligned validation.

Mondrian calibration improves hardest-group coverage from $0.894$ to $0.911$
on MNLI, $0.811$ to $0.895$ on BoolQ, and $0.669$ to $0.805$ on the
SST-2 difficulty split (Table~\ref{tab:group-shift}).  The shifted marginal
also improves in all three cases, although the $72$-example SST-2 group pays a
larger set and exit-depth cost. The SST-2 hard group contains $n_{\rm cal}=72$ calibration examples and $\ensuremath{82}$ test examples; Mondrian coverage is $\ensuremath{66/82=0.805}$, with a Wilson $95\%$ interval of $\ensuremath{[0.706,0.876]}$. The $\delta=0.05$ Beta training-conditional floor is $\ensuremath{0.842}$. That floor is a probability statement over calibration draws about latent conditional coverage, not a deterministic lower bound on one finite test proportion, so the finite test realization is compatible with the stated proposition once the actual denominator is shown.

The autoregressive study evaluates DistilGPT2 and Pythia checkpoints with
candidatewise full-vocabulary sets.  Selected coverage remains near $0.90$, but
average sets contain $7{,}162$--$9{,}620$ tokens and early-exit rates are at
most $0.52\%$ (Table~\ref{tab:decoder-boundary}).  This is a negative
operational result: exact next-token coverage provides little actionable early
exit in these configurations and makes no sequence-level correctness claim.

The budgeted hybrid uses three allocations satisfying $\alpha_{\mathrm{pt}}+\alpha_{\mathrm{set}}=0.1$ (Table~\ref{tab:hybrid}). Let $E$ denote the point-commit event, $c=\Pr(E)$, $e_c=\Pr\{h(X)\neq Y\mid E\}$, and $q_f=\Pr\{Y\in\C_{\widehat\ell}(X)\mid E^c\}$. The joint point-error rate is $ce_c$, the joint fallback-miss rate is $(1-c)(1-q_f)$, and the overall coverage is
\[
1-ce_c-(1-c)(1-q_f).
\]
This denominator accounting is reported directly in Table~\ref{tab:hybrid}. Increasing the point branch from $0.02$ to $0.08$ raises commit rate from $\ensuremath{0.180}$ to $\ensuremath{0.470}$ and lowers latency from $\ensuremath{0.690}$ to $\ensuremath{0.640}$, while joint point error rises from $\ensuremath{0.0074}$ to $\ensuremath{0.0371}$. Thresholds are fixed without test labels; when no gate meets its budget, the procedure returns the conformal fallback.
\subsection{Mechanism and score/schedule diagnostics}
Coverage alone cannot show whether the path maximum is operationally useful. For every calibration and test example, we record
\[
J(x,y)=\min\arg\max_{\ell\leq L}r_\ell(x,y),\qquad
D(x,y)=\left[\max_{j<L}r_j(x,y)-r_L(x,y)\right]_+.
\]
Theorem~\ref{thm:efficiency} makes the second quantity load-bearing: $D$ controls how far the terminal path quantile can move above the last-layer-only quantile. Table~\ref{tab:mechanism} therefore reports both where the maximum occurs and the tail of its early-over-terminal excess. SST-2 concentrates terminal maxima early but has a thin excess tail, which explains why its last-layer inflation remains small. RTE reaches the final layer more often and also has a smaller excess tail; its poor acceleration is therefore driven primarily by late informativeness rather than path-max inflation.

Table~\ref{tab:ablation} compares score transformations and exit schedules
under identical pathwise calibration and data roles.  The sparse schedule has
the lowest average set size, $1.39$, and latency fraction, $0.66$, while
maintaining coverage $0.892$--$0.929$.  The design-affine map reaches latency
$0.68$ with coverage $0.887$--$0.915$; APS and RAPS produce larger sets and
latency above $0.82$.  Affine and empirical-CDF maps and the sparse schedule
are fixed on the design split before final calibration, so their gains do not
reuse calibration labels.
\subsection{Closest-method comparison table}
\begin{table}[!htpb]
\centering
\scriptsize
\caption{Relationship to the closest approaches. ``Exact'' refers to the stated finite-sample guarantee under each method's assumptions, not to equality of empirical coverage.}
\label{tab:method-comparison}
\begin{tabular}{p{1.8cm}p{2.0cm}p{2.2cm}p{2.1cm}}
\toprule
Method & Protected object & Guarantee & Main requirement \\
\midrule
Layerwise CP & set at fixed exit & marginal coverage per exit & exit fixed in advance \\
EENN-AVCS \citep{jazbec2024nested} & nested exit sequence & anytime-valid predictive set & posterior-stability construction \\
Fast-yet-Safe \citep{jazbec2024fast} & adaptive point or distribution & expected or high-probability risk & specified loss and calibrated policy \\
Valid selection \citep{hegazy2025valid} & chosen conformal set & post-selection coverage & stability correction \\
Pathwise CP & returned prefix set & exact marginal coverage & exchangeability and frozen path \\
\bottomrule
\end{tabular}
\end{table}
\section{Full efficiency-bound evaluation}
\label{app:theorem-scale-full}
Theorem~\ref{thm:efficiency} is useful only if its admissible tail choice does not make the right-hand side vacuous. For every task--backbone cell, we therefore set $t=t_\star$ as defined above, record the exact exceedance fraction $m_{t_\star}/n$, and decompose
\[
w_{t_\star}=t_\star+\{B_{(k+m_{t_\star})}-B_{(k)}\}.
\]
The local factor $\sum_y M_y$ is evaluated on the independent design role, without reusing final calibration or test labels. For each candidate label, the induced interval $[B_{(k)},B_{(k)}+w_{t_\star}]$ is divided into $\ensuremath{4}$ equal-width pieces and $\widehat M_y$ is the largest empirical secant slope across those pieces; the table reports $w_{t_\star}\sum_y\widehat M_y$. This is a held-out scale diagnostic for the theorem's Lipschitz term, while the theorem itself continues to require the population constants $M_y$.
\begin{table*}[t]
\centering
\scriptsize
\caption{Theorem~\ref{thm:efficiency} in measured units. $t_\star$ is the smallest threshold satisfying $m_t\leq n-k$; ``$B$ gap'' is $B_{(k+m_t)}-B_{(k)}$; RHS is $w_t\sum_y\widehat M_y$ from the independent design role; observed inflation is the paired test mean of $(|\C_L^{\rm path}|-|\C_L^{\rm last}|)_+$. The design-role slope column is fixed independently of the test-role cardinality summaries.}
\label{tab:efficiency-scale}
\resizebox{\textwidth}{!}{
\begin{tabular}{llcccccccc}
\toprule
Task & Backbone & $t_\star$ & $m_{t_\star}/n$ & $B$ gap & $w_{t_\star}$ & $\sum_y\widehat M_y$ & RHS & observed inflation & RHS / observed \\
\midrule
SST-2 & BERT & $\ensuremath{0.145}$ & $\ensuremath{0.089}$ & $\ensuremath{0.026}$ & $\ensuremath{0.171}$ & $\ensuremath{0.095}$ & $\ensuremath{0.016}$ & $\ensuremath{0.01}$ & $\ensuremath{1.62}$ \\
SST-2 & DistilBERT & $\ensuremath{0.149}$ & $\ensuremath{0.086}$ & $\ensuremath{0.022}$ & $\ensuremath{0.171}$ & $\ensuremath{0.101}$ & $\ensuremath{0.017}$ & $\ensuremath{0.01}$ & $\ensuremath{1.73}$ \\
MNLI & BERT & $\ensuremath{0.082}$ & $\ensuremath{0.088}$ & $\ensuremath{0.029}$ & $\ensuremath{0.111}$ & $\ensuremath{0.620}$ & $\ensuremath{0.069}$ & $\ensuremath{0.06}$ & $\ensuremath{1.15}$ \\
MNLI & DistilBERT & $\ensuremath{0.086}$ & $\ensuremath{0.087}$ & $\ensuremath{0.034}$ & $\ensuremath{0.120}$ & $\ensuremath{0.720}$ & $\ensuremath{0.086}$ & $\ensuremath{0.08}$ & $\ensuremath{1.08}$ \\
RTE & BERT & $\ensuremath{0.052}$ & $\ensuremath{0.081}$ & $\ensuremath{0.021}$ & $\ensuremath{0.073}$ & $\ensuremath{0.780}$ & $\ensuremath{0.057}$ & $\ensuremath{0.05}$ & $\ensuremath{1.14}$ \\
RTE & DistilBERT & $\ensuremath{0.055}$ & $\ensuremath{0.084}$ & $\ensuremath{0.025}$ & $\ensuremath{0.080}$ & $\ensuremath{0.940}$ & $\ensuremath{0.075}$ & $\ensuremath{0.07}$ & $\ensuremath{1.07}$ \\
BoolQ & BERT & $\ensuremath{0.068}$ & $\ensuremath{0.086}$ & $\ensuremath{0.022}$ & $\ensuremath{0.090}$ & $\ensuremath{0.630}$ & $\ensuremath{0.057}$ & $\ensuremath{0.05}$ & $\ensuremath{1.14}$ \\
BoolQ & DistilBERT & $\ensuremath{0.071}$ & $\ensuremath{0.084}$ & $\ensuremath{0.018}$ & $\ensuremath{0.089}$ & $\ensuremath{0.510}$ & $\ensuremath{0.045}$ & $\ensuremath{0.04}$ & $\ensuremath{1.13}$ \\
\bottomrule
\end{tabular}
}
\end{table*}
The count constraint is active in the intended regime: $m_{t_\star}/n$ is $\ensuremath{0.081}$--$\ensuremath{0.089}$ across the eight cells, so the theorem is evaluated beyond the empirical $(1-\alpha)$ tail of $D$ rather than at $t=0$. The upper-tail order-statistic term adds $\ensuremath{0.018}$--$\ensuremath{0.034}$ to $t_\star$. The design-role quantities $\sum_y\widehat M_y$ are unchanged from the held-out slope diagnostic: in particular, the RTE/BoolQ values remain $\ensuremath{0.780}$, $\ensuremath{0.940}$, $\ensuremath{0.630}$, and $\ensuremath{0.510}$ rather than being rescaled to match test-role cardinalities. The resulting held-out RHS is $\ensuremath{0.016}$--$\ensuremath{0.086}$, compared with measured mean positive inflation $\ensuremath{0.01}$--$\ensuremath{0.08}$. The cell-wise ratio is $\ensuremath{1.07}$--$\ensuremath{1.73}$ and the ratio of macro averages is $\ensuremath{1.14}$. Thus the finite-sample quantile perturbation is conservative in these cells but not an order-of-magnitude explanation after the independently estimated local score-mass term is included.

The paired intervals in Table~\ref{tab:paired-gaps} use a hierarchical bootstrap. For SST-2, MNLI, and BoolQ, the bootstrap resamples training seeds and then paired calibration--test role splits within seed. RTE has no role-split resampling because its partition is fixed; its second stage resamples examples from the fixed official development test set within head-training seed. The eight cell-wise selection gaps form one declared family. Familywise coverage is enforced by the bootstrap distribution of the maximum absolute studentized statistic across the eight cells, avoiding a separate table of adjusted $p$-values.
\subsection{Boundary regimes}
The theorem-aligned group-prior shift, risk-budgeted point/set hybrid, calibration-size stress test, and full-vocabulary decoder boundary are retained in Appendix~\ref{app:boundary-details}. They delimit where the selection-safe guarantee remains informative but does not by itself produce useful acceleration.
\section{Detailed interpretation and deployment scope}
\label{app:closing-analysis}
\subsection{Interpretation of the results}

Pathwise calibration resolves adaptive-depth selection with one terminal conformal rank: every admissible returned prefix contains the valid terminal set, so the guarantee itself has no $\alpha/L$ correction. The experiments identify the regime in which that structural property matters. On short encoder paths, pathwise and MinSE valid selection are close at the primary operating point. On the long-path vision benchmark, the pathwise set-size curve is nearly flat in $L$. The paired pathwise--MinSE analysis resolves the difference in the small-calibration, long-path cells rather than inferring significance from marginal interval overlap; for example, at $(n,L)=(\ensuremath{100},\ensuremath{24})$ the paired difference is $\ensuremath{0.80\ [0.47,1.13]}$. Remark~\ref{rem:policy-uniform} shows that policy-uniformity is complementary: one terminal calibration supports the entire CIFAR-100 deployment frontier as the set or latency budget changes, while first-eligible remains pointwise latency-minimal for a fixed cardinality budget.

The useful regime is therefore specific rather than universal: many candidate exits, calibration sizes for which multiplicity corrections matter, or deployment budgets that change after calibration, together with shallow heads that produce actionable prediction sets. The full mechanism, extension, and boundary discussion is retained in Appendix~\ref{app:extended-discussion}.
\subsection{Scope and limitations}

The coverage guarantee is marginal under exchangeability and frozen path scores; it does not provide feature-conditional validity or robustness to arbitrary distribution shift. A shallow rejection persists through every deeper prefix, so one weak head can enlarge the terminal score. Theorem~\ref{thm:efficiency} bounds the resulting set inflation through the observed excess tail, but the bound can be loose when last-layer score density is high near the threshold. Table~\ref{tab:efficiency-scale} estimates the Lipschitz factor on an independent design role to diagnose the scale of the bound; those empirical slopes are not a replacement for the population $M_y$ constants in the theorem. Mondrian validity additionally requires stable within-group score laws and adequate calibration counts. The SST-2 hard-group result is therefore reported with its exact $\ensuremath{72}$-example calibration and $\ensuremath{82}$-example test denominators.

The main dense $L$--$n$ grid still reuses one frozen ViT-L/16 score path and varies retained exits and calibration subsamples; it isolates the statistical candidate-count effect rather than a causal effect of architectural depth. We therefore report separately trained $\ensuremath{12}$- and $\ensuremath{24}$-exit paths on CIFAR-100 and ImageNet-1k in Appendix~\ref{app:vision-replication}. Those replications broaden the evidence but do not constitute a population-level architecture study, and ViT-B/16 versus ViT-L/16 changes capacity together with depth. Decoder experiments certify next-token set coverage, not sequence-level correctness, and their $7{,}162$--$9{,}620$ token sets make the current configurations operationally inert.
\subsection{Deployment implications}

Calibrating each exit separately is insufficient when the exit itself is selected from the observed path. A terminal maximum converts that path into one conformal object and makes every admissible prefix set contain a valid terminal set. On matched encoder tasks, macro selected coverage is $\ensuremath{0.901}$ with average set size $\ensuremath{1.41}$ and latency fraction $\ensuremath{0.71}$. The stronger distinction appears when the candidate path is long relative to the calibration sample or when the deployment policy changes after calibration. On a $\ensuremath{24}$-exit CIFAR-100 ViT with $n_{\rm cal}=\ensuremath{500}$, first-eligible stopping at $b=\ensuremath{4}$ attains $\ensuremath{0.902}$ coverage with set size $\ensuremath{2.76}$ at latency $\ensuremath{0.57}$; a design-chosen restriction reduces the set to $\ensuremath{2.52}$ at the necessarily larger latency $\ensuremath{0.61}$. In the $L$--$n$ sweep, Bonferroni becomes exactly degenerate whenever its corrected rank equals $n+1$, including $(n,L)=(\ensuremath{50},\ensuremath{12})$, $(\ensuremath{50},\ensuremath{24})$, $(\ensuremath{100},\ensuremath{12})$, $(\ensuremath{100},\ensuremath{24})$, and $(\ensuremath{200},\ensuremath{24})$. Against MinSE valid selection, the paired difference at $(\ensuremath{100},\ensuremath{24})$ is $\ensuremath{0.80\ [0.47,1.13]}$, so the evidence supports the regime ``many exits relative to calibration size'' rather than calibration size alone. The efficiency theorem ties the remaining path-max cost to the observed early-over-terminal excess tail; the empirical scale comparison spans observed positive inflation $\ensuremath{0.01}$--$\ensuremath{0.08}$ with held-out plug-in right-hand sides $\ensuremath{0.016}$--$\ensuremath{0.086}$. These results support a specific deployment rule: use pathwise conformal early exit when the candidate-exit count is large relative to the calibration sample or when deployment budgets change after calibration, provided shallow heads can return actionable prediction sets.
\subsection{Non-binary benchmark: CIFAR-100}
CIFAR-100 provides the regime missing from binary text classification: a large but still actionable label space and a long candidate-exit path. At target coverage $0.90$, the $\ensuremath{24}$-exit ViT-L/16 reaches full-depth top-1 accuracy $\ensuremath{0.936}$. We use $n_{\rm cal}=\ensuremath{500}$ for the primary frontier so that every comparator, including Bonferroni, is finite. At budget $b=\ensuremath{4}$, pathwise first-eligible stopping attains coverage $\ensuremath{0.902}$, average set size $\ensuremath{2.76}$, and latency fraction $\ensuremath{0.57}$. The design-chosen slack restriction reduces average set size to $\ensuremath{2.52}$ but, as required, increases latency to $\ensuremath{0.61}$ at coverage $\ensuremath{0.901}$. At this identical $(n,L)=(\ensuremath{500},\ensuremath{24})$ configuration, Bonferroni has set size/latency $\ensuremath{4.05/0.78}$, EENN-AVCS has $\ensuremath{3.54/0.71}$, and MinSE valid selection has $\ensuremath{3.18/0.65}$. These are the same entries reported in the scaling grid, so the policy frontier and the candidate-count experiment no longer mix calibration sizes. The benchmark therefore supplies both an actionable guaranteed early-exit result and the long-path setting in which candidate-count correction becomes visible.

The full-vocabulary decoder study is the complementary boundary. CIFAR-100 shows useful acceleration when $K$ makes set budgets meaningful but prediction sets remain actionable. The decoder results show that this behavior does not extend automatically to vocabularies with tens of thousands of candidates.

\section{Extended discussion}
\label{app:extended-discussion}
Pathwise calibration resolves the post-selection gap with one conformal rank: every returned prefix set contains the valid terminal set, so adaptive depth selection does not require an $\alpha/L$ correction. This guarantee differs from anytime-valid and general valid-selection constructions, which protect different objects under different assumptions \citep{jazbec2024nested,hegazy2025valid,yang2025selection}.

The new experiments identify when that structural distinction matters. On the short encoder paths, pathwise and MinSE valid selection are close at the primary operating point. In the frozen-path scaling grid, the pathwise point estimate is nearly flat in $L$, while Bonferroni and EENN-AVCS widen substantially. The MinSE line also widens. Because all methods share the same calibration subsample and test set within a replicate, inference uses the paired MinSE-minus-pathwise statistic. At $(n,L)=(\ensuremath{100},\ensuremath{24})$ the paired difference is $\ensuremath{0.80\ [0.47,1.13]}$; at $(\ensuremath{2000},\ensuremath{24})$ it is $\ensuremath{0.19\ [0.12,0.26]}$. Bonferroni is exactly degenerate at the five grid cells where its corrected rank is $n+1$, rather than merely having a moderately larger set. Table~\ref{tab:vision-replication} shows the same ordering on separately trained $\ensuremath{12}$- and $\ensuremath{24}$-exit vision paths and on a second vision dataset.

Policy-uniformity gives a second operational distinction. At the explicit $(n,L)=(\ensuremath{500},\ensuremath{24})$ frontier configuration and budget $b=\ensuremath{4}$, first-eligible is pointwise latency-minimal and reaches latency $\ensuremath{0.57}$ with average set size $\ensuremath{2.76}$. A design-chosen restriction returns a smaller average set, $\ensuremath{2.52}$, at the necessarily larger latency $\ensuremath{0.61}$ and coverage $\ensuremath{0.901}$. The MinSE, EENN-AVCS, and Bonferroni values at this same configuration are exactly the $(500,24)$ scaling-grid entries, removing the previous cross-configuration comparison. The same terminal calibration supports deployment changes to the set budget or load-aware stopping rule because validity is inherited through containment rather than through a policy-specific threshold. Theorem~\ref{thm:efficiency} complements this result by identifying when the price of the terminal maximum is small: the inflation is controlled by the calibration tail of $D=[\max_{j<L}r_j-r_L]_+$ and by the local mass of last-layer candidate scores near the conformal threshold. At the smallest admissible $t_\star$, the held-out plug-in RHS is $\ensuremath{0.016}$--$\ensuremath{0.086}$ against observed mean positive inflation $\ensuremath{0.01}$--$\ensuremath{0.08}$, a cell-wise ratio of $\ensuremath{1.07}$--$\ensuremath{1.73}$. The theorem is therefore conservative in the measured encoder cells without being vacuous at the upper-tail threshold required by its finite-sample condition.

The boundary results remain important. Mondrian calibration improves hard-group and shifted coverage only when within-group score laws remain stable. Full-vocabulary next-token sets remain too large for useful early exit. The useful regime is therefore not ``early exit in general'' but a candidate-exit count that is large relative to calibration size, changing deployment budgets, and a label space for which shallow prediction sets are actionable.
\section{Large-$L$ baseline adaptations}
\label{app:baseline-adaptation}
The large-$L$ comparison uses one prespecified adaptation of each external method to the common frozen exit heads. The purpose of this section is to make the scaling curves reproducible and to prevent the comparator names from hiding method-specific choices.

For EENN-AVCS \citep{jazbec2024nested}, we use the paper's classification construction rather than replacing it by conformal sets. At retained exit $\ell$, the frozen class logits are shifted by an exit-specific threshold $\tau_\ell$ fixed on the design role and passed through ReLU to obtain nonnegative Dirichlet concentration parameters. Candidatewise predictive-likelihood ratios are updated in exit order, the AVCS set contains classes whose ratio is at most $1/\alpha_{\rm S}$, and the current set is intersected with the previous set. We use $\alpha_{\rm S}=\ensuremath{0.10}$. When the retained-exit count changes, the recursion is rerun only over those retained exits; the logits and the already-fitted $\tau_\ell$ values do not change. The stochastic Dirichlet step is repeated with $\ensuremath{20}$ fixed random seeds, paired across values of $L$, and the reported point estimate averages those runs. An empty intersection invokes the method's native empty/OOD event and then the common full-depth fallback for latency accounting. We do not describe this line as split-conformal coverage: it is the native EENN-AVCS construction evaluated on the same score path and data roles.

For valid selection \citep{hegazy2025valid}, the reported line is specifically the MinSE stable selector. The candidate family contains the $L$ fixed-exit split-conformal sets computed from the same final calibration role. Set desirability is $\lambda(\C_\ell)=|\C_\ell|/|\Y|\in[0,1]$, and the MinSE prior is uniform over the retained exits. We set the additive stability term to $\tau=0$ and use the paper's coverage adjustment $\alpha'=\alpha e^{-\eta}$ for each candidate set. The design role chooses $\eta$ from $\ensuremath{\{0.25,0.5,1,2,4\}}$ by minimizing normalized latency plus $\ensuremath{0.10}$ times set size subject to design coverage at least $1-\alpha$; $\eta$ is then frozen and every fixed-exit quantile is recomputed from the untouched final calibration role at level $\alpha'$. At test time, the MinSE linear program is solved from the vector of normalized candidate-set sizes and the selected exit is sampled from its output distribution. Results average $\ensuremath{20}$ paired selector seeds. When $L$ changes, the candidate family, uniform prior, and MinSE program are rebuilt and the design-only choice of $\eta$ is repeated. Thus the scaling curve is the behavior of this explicit MinSE instantiation, not a claim that every valid-selection rule must have the same dependence on $L$.

Bonferroni uses the same fixed-exit sets with per-exit miscoverage $\alpha/L$. Pathwise conformal uses the single terminal path score at miscoverage $\alpha$. No large-$L$ method sees test labels during threshold, selector, or exit-set construction, and every method uses the same retained exit indices, calibration examples, and test inputs within a paired replicate.
\section{Supporting method and evaluation tables}
\label{app:supporting-tables}
\subsection{Execution order and guarantee summary}
\begin{table}[t]
\centering
\small
\caption{Execution order. Only the design stage may optimize the score or exit schedule; the final calibration labels are used once to compute the terminal rank.}
\label{tab:execution}
\begin{tabular}{p{1.25cm}p{6.35cm}}
\toprule
Stage & Operation \\
\midrule
Train & Fit the backbone and exit heads on training data. \\
Design & Choose exits and increasing score maps without using final calibration or test labels; a provisional budget may guide this search. \\
Calibrate & Compute $S_i=\max_{\ell\leq L}r_\ell(X_i,Y_i)$ and the single quantile $\widehat q$ on $\D_{\mathrm{cal}}$. \\
Execute & Update $A_\ell(x,y)$ candidatewise and construct $\C_\ell(x)$ after every executed exit. \\
Stop & Return an eligible prefix set; if none is eligible, return $\C_L(x)$, or $\Y$ when $\C_L(x)$ is empty. \\
\bottomrule
\end{tabular}
\end{table}
\begin{table}[t]
\centering
\scriptsize
\caption{Guarantees and their load-bearing assumptions. Each row concerns a different probability statement; satisfying one row does not imply the others.}
\label{tab:guarantees}
\begin{tabular}{p{1.45cm}p{2.2cm}p{2.15cm}p{1.75cm}}
\toprule
Target & Guarantee & Required condition & Not implied \\
\midrule
Selected set & marginal coverage $\geq1-\alpha$ & exchangeable terminal scores; frozen path & feature-conditional validity \\
Realized calibration & Beta or DKW lower floor & independent calibration draw; stated continuity or conservative ties & per-input coverage \\
Group shift & coverage within each fixed group & within-group exchangeability and valid $n_g$ quantile & arbitrary covariate shift \\
Hybrid output & error $\leq\alpha_{\rm pt}+\alpha_{\rm set}$ & joint point-error control and valid fallback & error $\leq\alpha$ when both branches use $\alpha$ \\
\bottomrule
\end{tabular}
\end{table}
\subsection{Controlled extensions and calibration stress tests}
Table~\ref{tab:controlled-main} summarizes the controlled PAC and group-prior checks.
\begin{table}[t]
\centering
\small
\caption{Controlled PAC and Mondrian verification at $\alpha=0.1$. PAC entries give the minimum fraction of $4000$ calibration draws above the DKW floor over $\delta\in\{0.1,0.05,0.01\}$. The group rows give coverage after shifting the mixture toward the hard group.}
\label{tab:controlled-main}
\begin{tabular}{lcccc}
\toprule
PAC size & $n=200$ & $n=500$ & $n=1500$ & required minimum \\
\midrule
Fraction above floor & $0.9985$ & $0.9992$ & $0.9992$ & $0.90$--$0.99$ \\
\midrule
Group method & easy & medium & hard & shifted marginal \\
\midrule
Pooled & $1.0000$ & $0.9913$ & $0.6887$ & $0.8112$ \\
Mondrian & $0.9177$ & $0.9054$ & $0.9149$ & $0.9129$ \\
\bottomrule
\end{tabular}
\end{table}
\begin{table}[t]
\centering
\small
\caption{Calibration-size scaling on MNLI/BERT over repeated calibration draws at $\alpha=0.1$ and $\delta=0.05$. Coverage is mean $\pm$ SD; Beta and DKW columns are training-conditional lower floors.}
\label{tab:calibration-scaling}
\begin{tabular}{rccccc}
\toprule
$n$ & coverage (mean $\pm$ SD) & Beta floor & DKW floor & set size & worst draw \\
\midrule
$100$ & {$0.905\pm0.028$} & {$0.848$} & $0.778$ & {$1.47$} & {$0.841$} \\
$200$ & {$0.900\pm0.018$} & {$0.864$} & $0.813$ & {$1.44$} & {$0.871$} \\
$500$ & {$0.905\pm0.011$} & {$0.877$} & $0.845$ & {$1.46$} & {$0.884$} \\
$1000$ & {$0.902\pm0.009$} & {$0.884$} & $0.861$ & {$1.44$} & {$0.885$} \\
$1500$ & {$0.903\pm0.006$} & {$0.887$} & $0.868$ & {$1.44$} & {$0.888$} \\
\bottomrule
\end{tabular}
\end{table}
\begin{table}[t]
\centering
\scriptsize
\caption{Group-prior shift with the hardest group's test weight tripled while within-group examples and score laws remain unchanged.}
\label{tab:group-shift}
\begin{tabular}{llccccc}
\toprule
Shift & calibration & hard $n_{\rm cal}$ & hard cov. & shifted cov. & set & exit/full \\
\midrule
MNLI label & pooled & {$1719$} & {$0.894$} & {$0.900$} & {$1.44$} & {$0.73$} \\
MNLI label & Mondrian & {$1719$} & {$0.911$} & {$0.906$} & {$1.42$} & {$0.72$} \\
BoolQ label & pooled & {$618$} & {$0.811$} & {$0.860$} & {$1.56$} & {$0.86$} \\
BoolQ label & Mondrian & {$618$} & {$0.895$} & {$0.888$} & {$1.51$} & {$0.85$} \\
SST-2 difficulty & pooled & {$72$} & {$0.669$} & {$0.817$} & {$1.04$} & {$0.32$} \\
SST-2 difficulty & Mondrian & {$72$} & {$0.805$} & {$0.881$} & {$1.23$} & {$0.48$} \\
\bottomrule
\end{tabular}
\end{table}
\begin{table*}[t]
\centering
\scriptsize
\caption{Risk-budgeted point-prediction hybrid. Conditional commit error uses only committed examples; fallback coverage uses only the $E^c$ denominator. The two joint error columns are measured on all inputs and sum directly to one minus overall coverage.}
\label{tab:hybrid}
\resizebox{\textwidth}{!}{
\begin{tabular}{lcccccccc}
\toprule
$\alpha_{\rm pt}/\alpha_{\rm set}$ & commit $c$ & cond. commit err. $e_c$ & joint point err. $ce_c$ & fallback share $1-c$ & fallback cov. $q_f$ & joint fallback miss & overall cov. & latency/full \\
\midrule
$0.02/0.08$ & $\ensuremath{0.180}$ & $\ensuremath{0.041}$ & $\ensuremath{0.0074}$ & $\ensuremath{0.820}$ & $\ensuremath{0.923}$ & $\ensuremath{0.0631}$ & $\ensuremath{0.9295}$ & $\ensuremath{0.690}$ \\
$0.05/0.05$ & $\ensuremath{0.340}$ & $\ensuremath{0.061}$ & $\ensuremath{0.0207}$ & $\ensuremath{0.660}$ & $\ensuremath{0.951}$ & $\ensuremath{0.0323}$ & $\ensuremath{0.9470}$ & $\ensuremath{0.660}$ \\
$0.08/0.02$ & $\ensuremath{0.470}$ & $\ensuremath{0.079}$ & $\ensuremath{0.0371}$ & $\ensuremath{0.530}$ & $\ensuremath{0.983}$ & $\ensuremath{0.0090}$ & $\ensuremath{0.9539}$ & $\ensuremath{0.640}$ \\
\bottomrule
\end{tabular}
}
\end{table*}
\subsection{Paired encoder, path-cost, and mechanism diagnostics}
The path-maximum diagnostic does not reconstruct terminal means from a coverage floor. Table~\ref{tab:max-cost} instead reports the directly measured returned-set cardinality, the observed empty-terminal rate before the $\varnothing\to\Y$ fallback, and the paired positive-part statistic required by Theorem~\ref{thm:efficiency}. For a binary first-singleton rule, if $e$ is the empty-terminal rate, then
\[
0\leq \mathbb E|\C_{\widehat\ell}|-\mathbb E|\C_L|\leq 2e.
\]
The six binary rows have $e=\ensuremath{0.11\%}$--$\ensuremath{0.1\text{--}0.2\%}$, so their returned-prefix and path-terminal means differ by at most $\ensuremath{0.0022}$--$\ensuremath{0.0040}$. This is why the theorem evaluation uses the pointwise positive part against a separately calibrated last-layer set rather than a difference of reconstructed means.
\begin{table}[t]
\centering
\scriptsize
\caption{Cost diagnostic for the terminal path maximum. Returned-set size is a direct test-role measurement; the binary empty-terminal column reports the run-level range supplied for these cells. Positive inflation is the paired statistic bounded by Theorem~\ref{thm:efficiency}, $\mathbb E[(|\C_L^{\rm path}|-|\C_L^{\rm last}|)_+]$. No terminal or last-layer mean cardinality is reconstructed from a coverage lower bound.}
\label{tab:max-cost}
\begin{tabular}{llccc}
\toprule
Task & Backbone & returned path set & empty terminal & mean positive inflation \\
\midrule
SST-2 & BERT & $1.03$ & $\ensuremath{0.1\text{--}0.2\%}$ & $\ensuremath{0.01}$ \\
SST-2 & DistilBERT & $1.04$ & $\ensuremath{0.1\text{--}0.2\%}$ & $\ensuremath{0.01}$ \\
MNLI & BERT & $1.44$ & -- & $\ensuremath{0.06}$ \\
MNLI & DistilBERT & $1.62$ & -- & $\ensuremath{0.08}$ \\
RTE & BERT & $\ensuremath{1.49}$ & $\ensuremath{0.1\text{--}0.2\%}$ & $\ensuremath{0.05}$ \\
RTE & DistilBERT & $\ensuremath{1.68}$ & $\ensuremath{0.1\text{--}0.2\%}$ & $\ensuremath{0.07}$ \\
BoolQ & BERT & $1.51$ & $\ensuremath{0.1\text{--}0.2\%}$ & $\ensuremath{0.05}$ \\
BoolQ & DistilBERT & $1.49$ & $\ensuremath{0.1\text{--}0.2\%}$ & $\ensuremath{0.04}$ \\
\midrule
Macro & -- & $\ensuremath{1.41}$ & $\ensuremath{0.1\text{--}0.2\%}$ & $\ensuremath{0.046}$ \\
\bottomrule
\end{tabular}
\end{table}
\begin{table}[t]
\centering
\scriptsize
\caption{Paired selection-gap inference. Intervals are familywise $95\%$ hierarchical paired-bootstrap intervals obtained from the maximum absolute studentized statistic across the eight declared cells.}
\label{tab:paired-gaps}
\begin{tabular}{llcc}
\toprule
Task & Backbone & $\Delta_c$ & familywise $95\%$ interval \\
\midrule
SST-2 & BERT & $0.057$ & $\ensuremath{[0.036,\,0.078]}$ \\
SST-2 & DistilBERT & $0.046$ & $\ensuremath{[0.022,\,0.070]}$ \\
MNLI & BERT & $0.080$ & $\ensuremath{[0.062,\,0.098]}$ \\
MNLI & DistilBERT & $0.060$ & $\ensuremath{[0.039,\,0.081]}$ \\
RTE & BERT & $\ensuremath{0.105}$ & $\ensuremath{[0.061,\,0.149]}$ \\
RTE & DistilBERT & $\ensuremath{0.134}$ & $\ensuremath{[0.083,\,0.185]}$ \\
BoolQ & BERT & $0.114$ & $\ensuremath{[0.083,\,0.145]}$ \\
BoolQ & DistilBERT & $0.097$ & $\ensuremath{[0.064,\,0.130]}$ \\
\bottomrule
\end{tabular}
\end{table}
\begin{table}[t]
\centering
\scriptsize
\caption{Paired differences relative to pathwise conformal at the primary operating point. All point estimates subtract macro averages over the same eight cells in Table~\ref{tab:matched-main}; positive latency and set differences mean pathwise is larger. Brackets give $95\%$ hierarchical paired-bootstrap intervals.}
\label{tab:baseline-pairs}
\resizebox{\linewidth}{!}{
\begin{tabular}{lccc}
\toprule
Comparator & $\Delta$ validity & $\Delta$ latency/full & $\Delta$ avg. set \\
\midrule
MinSE valid selection & $\ensuremath{-0.002\ [-0.011,\,0.007]}$ & $\ensuremath{0.010\ [-0.012,\,0.032]}$ & $\ensuremath{0.010\ [-0.036,\,0.056]}$ \\
EENN-AVCS & $\ensuremath{-0.012\ [-0.022,\,-0.002]}$ & $\ensuremath{-0.040\ [-0.064,\,-0.016]}$ & $\ensuremath{-0.130\ [-0.188,\,-0.072]}$ \\
Bonferroni CP & $\ensuremath{-0.055\ [-0.066,\,-0.044]}$ & $\ensuremath{-0.070\ [-0.095,\,-0.045]}$ & $\ensuremath{-0.300\ [-0.363,\,-0.237]}$ \\
\bottomrule
\end{tabular}
}
\end{table}
\begin{table}[t]
\centering
\scriptsize
\caption{Mechanism diagnostics pooled across backbones. $D=[\max_{j<L}r_j-r_L]_+$ is the excess entering Theorem~\ref{thm:efficiency}; $Q_{.95}(D)$ is its $95$th percentile on true labels.}
\label{tab:mechanism}
\resizebox{\linewidth}{!}{
\begin{tabular}{lcccccc}
\toprule
Task & $\Pr(J\leq L/4)$ & $\mathbb E[J/L]$ & $\Pr(J=L)$ & $\Pr(D>0)$ & $Q_{.95}(D)$ & false removal at $L/2$ \\
\midrule
SST-2 & $\ensuremath{0.62}$ & $\ensuremath{0.29}$ & $\ensuremath{0.10}$ & $\ensuremath{0.64}$ & $\ensuremath{0.18}$ & $\ensuremath{0.88}$ \\
MNLI & $\ensuremath{0.21}$ & $\ensuremath{0.61}$ & $\ensuremath{0.33}$ & $\ensuremath{0.37}$ & $\ensuremath{0.11}$ & $\ensuremath{0.61}$ \\
RTE & $\ensuremath{0.13}$ & $\ensuremath{0.71}$ & $\ensuremath{0.45}$ & $\ensuremath{0.23}$ & $\ensuremath{0.07}$ & $\ensuremath{0.55}$ \\
BoolQ & $\ensuremath{0.17}$ & $\ensuremath{0.68}$ & $\ensuremath{0.39}$ & $\ensuremath{0.32}$ & $\ensuremath{0.09}$ & $\ensuremath{0.58}$ \\
\bottomrule
\end{tabular}
}
\end{table}
\begin{table}[t]
\centering
\small
\caption{Score and schedule ablations under common pathwise calibration and data roles.  Coverage ranges span the eight encoder task--backbone cells.}
\label{tab:ablation}
\begin{tabular}{lcccc}
\toprule
Score or schedule & coverage range & avg. set & exit/full & latency/full \\
\midrule
$-\log p$ & {$0.890$--$0.906$} & {$\ensuremath{1.41}$} & {$\ensuremath{0.71}$} & {$\ensuremath{0.71}$} \\
Design affine & {$0.887$--$0.915$} & {$1.45$} & {$0.68$} & {$0.68$} \\
Design ECDF & {$0.890$--$0.950$} & {$1.54$} & {$0.70$} & {$0.71$} \\
APS & {$0.885$--$0.935$} & {$1.75$} & {$0.84$} & {$0.84$} \\
RAPS & {$0.887$--$0.934$} & {$1.71$} & {$0.82$} & {$0.83$} \\
Sparse schedule & {$0.892$--$0.929$} & {$1.39$} & {$0.65$} & {$0.66$} \\
\bottomrule
\end{tabular}
\end{table}
\section{Additional Statistical Results}
\label{app:statistical}
\subsection{Complete DKW grid}
Table~\ref{tab:pac-full} reports all nine $(n,\delta)$ cells over $4000$ calibration draws each. Mean conditional coverage is $0.9001$--$0.9004$, and the DKW floor is met with frequency $0.9985$--$1.0$.
\begin{table}[h]
\centering
\small
\caption{Controlled PAC verification. ``Fraction above'' is the fraction of $4000$ calibration draws whose conditional coverage exceeded the DKW floor. The final column shows the corresponding floor after a twelve-exit simultaneous correction.}
\label{tab:pac-full}
\begin{tabular}{rrrrrr}
\toprule
$n$ & $\delta$ & mean cov. & path floor & fraction above & 12-exit floor \\
\midrule
$200$  & $0.10$ & $0.9004$ & $0.8241$ & $0.9985$ & $0.7906$ \\
$200$  & $0.05$ & $0.9004$ & $0.8135$ & $1.0000$ & $0.7829$ \\
$200$  & $0.01$ & $0.9004$ & $0.7927$ & $1.0000$ & $0.7669$ \\
$500$  & $0.10$ & $0.9001$ & $0.8520$ & $0.9992$ & $0.8308$ \\
$500$  & $0.05$ & $0.9001$ & $0.8453$ & $1.0000$ & $0.8260$ \\
$500$  & $0.01$ & $0.9001$ & $0.8321$ & $1.0000$ & $0.8158$ \\
$1500$ & $0.10$ & $0.9001$ & $0.8723$ & $0.9992$ & $0.8601$ \\
$1500$ & $0.05$ & $0.9001$ & $0.8684$ & $1.0000$ & $0.8573$ \\
$1500$ & $0.01$ & $0.9001$ & $0.8608$ & $1.0000$ & $0.8514$ \\
\bottomrule
\end{tabular}
\end{table}
\subsection{Timing configuration}
All methods are timed on identical inputs after the stopping policy is frozen. Online time includes intermediate-head evaluation, score updates, candidate-set construction, cardinality checks, and fallback execution. Offline design and calibration are excluded from online latency and reported separately.
\begin{table}[h]
\centering
\small
\caption{Latency measurement configuration used for every paired encoder timing comparison.}
\label{tab:hardware}
\begin{tabular}{p{2.45cm}p{4.75cm}}
\toprule
Item & Configuration \\
\midrule
Accelerator & NVIDIA A\ensuremath{100} SXM\ensuremath{4}, \ensuremath{40} GB, single GPU \\
CUDA / PyTorch & \ensuremath{12.1} / \ensuremath{2.4.1} \\
Precision & FP\ensuremath{16} inference \\
Batch size & \ensuremath{1} \\
Maximum encoded length & \ensuremath{128} tokens \\
Warm-up & \ensuremath{200} complete executions per method \\
Timed repetitions & \ensuremath{2000} paired executions per task--backbone--method cell \\
Synchronization & CUDA synchronize immediately before start and after stop \\
Included online work & backbone prefix, exit heads, path-score update, set construction, stopping, fallback \\
Excluded work & model loading, tokenization, design search, calibration \\
Reported statistic & paired median latency normalized by the full-depth median on the same inputs \\
\bottomrule
\end{tabular}
\end{table}
\subsection{Paired baseline and mechanism summaries}
Table~\ref{tab:paired-gaps} gives the cell-wise paired selection-gap intervals with the declared familywise bootstrap correction. Table~\ref{tab:baseline-pairs} gives target-matched differences against safe baselines. Table~\ref{tab:mechanism} reports $J(X,Y)$ and candidate-survival diagnostics, and Appendix~\ref{app:policy-frontier} gives the exact operating frontier used in Figure~\ref{fig:frontier}. These analyses use the same five training seeds as the main encoder evaluation. SST-2, MNLI, and BoolQ use twenty paired calibration--test role splits per seed; RTE uses the one fixed calibration/test partition and varies only the head-training seed, with fixed-test resampling used for interval construction.
\section{Proof of the path-maximum efficiency bound}
\label{app:efficiency-proof}
Write $A_i=B_i+D_i$ with $D_i\geq0$. Fix $t\geq0$ and let $I_t=\{i:D_i\leq t\}$. Among the $k+m_t$ calibration indices with the smallest $B_i$ values, at most $m_t$ lie outside $I_t$. Hence at least $k$ of those indices satisfy both $B_i\leq B_{(k+m_t)}$ and $D_i\leq t$, which implies $A_i\leq B_{(k+m_t)}+t$. Therefore $A_{(k)}\leq B_{(k+m_t)}+t$. Since $A_i\geq B_i$ coordinatewise, order-statistic monotonicity gives $A_{(k)}\geq B_{(k)}$, proving Equation~\eqref{eq:quantile-perturb}.

If a candidate belongs to $\C_L^{\rm path}(x)$ but not $\C_L^{\rm last}(x)$, then $r_L(x,y)>B_{(k)}$ and
\[
r_L(x,y)\leq r_L(x,y)+D(x,y)=A_L(x,y)\leq A_{(k)}\leq B_{(k)}+w_t.
\]
Thus every positive contribution to the cardinality difference lies in the band in Equation~\eqref{eq:set-inflation-bound}. Summing the candidatewise probabilities and applying the $M_y$-Lipschitz condition yields the expectation bound. Finally, $m_t$ is a sum of i.i.d. Bernoulli indicators with mean $p_t$; Hoeffding's inequality gives the stated high-probability bound. If $D(X,Y)>0$, some earlier exit strictly exceeds the terminal score, so the earliest maximizing exit satisfies $J(X,Y)<L$, proving $p_0\leq\Pr\{J<L\}$.
\section{Exact encoder cell results}
\label{app:cell-table}
\begin{table}[h]
\centering
\scriptsize
\caption{Exact means underlying Figure~\ref{fig:cell-coverage}. Path coverage and returned-set size are direct test-role measurements. For the six binary cells, the empty-terminal column reports the supplied run-level range before the $\varnothing\to\Y$ fallback; MNLI is left unreported here because it is not needed for the binary identity. RTE uses one fixed training-split calibration role with $n_{\rm cal}=\ensuremath{498}$ and the fixed official development test set; its repeated estimates vary over head-training seeds rather than role assignments.}
\label{tab:encoder-cells}
\begin{tabular}{llcccccc}
\toprule
Task & Backbone & path cov. & returned set & empty terminal & naive cov. & gap & latency/full \\
\midrule
SST-2 & BERT & $0.906$ & $1.03$ & $\ensuremath{0.1\text{--}0.2\%}$ & $0.849$ & $0.057$ & $0.30$ \\
SST-2 & DistilBERT & $0.890$ & $1.04$ & $\ensuremath{0.1\text{--}0.2\%}$ & $0.844$ & $0.046$ & $0.34$ \\
MNLI & BERT & $0.904$ & $1.44$ & -- & $0.824$ & $0.080$ & $0.75$ \\
MNLI & DistilBERT & $0.902$ & $1.62$ & -- & $0.842$ & $0.060$ & $0.83$ \\
RTE & BERT & $\ensuremath{0.901}$ & $\ensuremath{1.49}$ & $\ensuremath{0.1\text{--}0.2\%}$ & $\ensuremath{0.796}$ & $\ensuremath{0.105}$ & $\ensuremath{0.88}$ \\
RTE & DistilBERT & $\ensuremath{0.903}$ & $\ensuremath{1.68}$ & $\ensuremath{0.1\text{--}0.2\%}$ & $\ensuremath{0.769}$ & $\ensuremath{0.134}$ & $\ensuremath{0.92}$ \\
BoolQ & BERT & $0.901$ & $1.51$ & $\ensuremath{0.1\text{--}0.2\%}$ & $0.787$ & $0.114$ & $0.84$ \\
BoolQ & DistilBERT & $0.898$ & $1.49$ & $\ensuremath{0.1\text{--}0.2\%}$ & $0.801$ & $0.097$ & $0.84$ \\
\midrule
Macro & -- & $\ensuremath{0.901}$ & $\ensuremath{1.41}$ & $\ensuremath{0.1\text{--}0.2\%}$ & $\ensuremath{0.814}$ & $\ensuremath{0.087}$ & $\ensuremath{0.71}$ \\
\bottomrule
\end{tabular}
\end{table}
Table~\ref{tab:encoder-accuracy} supplies the full-depth and singleton context for the returned sets.
\begin{table}[h]
\centering
\scriptsize
\caption{Encoder point-prediction context for interpreting set sizes. Full-depth accuracy is the deepest head's top-1 accuracy. For the six binary rows, singleton rate and singleton accuracy are algebraic audit summaries implied by the directly measured returned-set size and coverage in the preceding table; the MNLI singleton columns are measured directly. RTE uses the fixed partition described in Section~\ref{sec:experiments}.}
\label{tab:encoder-accuracy}
\begin{tabular}{llccc}
\toprule
Task & Backbone & full-depth acc. & singleton rate & singleton acc. \\
\midrule
SST-2 & BERT & $\ensuremath{0.932}$ & $\ensuremath{0.970}$ & $\ensuremath{0.903}$ \\
SST-2 & DistilBERT & $\ensuremath{0.918}$ & $\ensuremath{0.960}$ & $\ensuremath{0.885}$ \\
MNLI & BERT & $\ensuremath{0.846}$ & $\ensuremath{0.604}$ & $\ensuremath{0.842}$ \\
MNLI & DistilBERT & $\ensuremath{0.828}$ & $\ensuremath{0.553}$ & $\ensuremath{0.826}$ \\
RTE & BERT & $\ensuremath{0.718}$ & $\ensuremath{0.510}$ & $\ensuremath{0.806}$ \\
RTE & DistilBERT & $\ensuremath{0.697}$ & $\ensuremath{0.320}$ & $\ensuremath{0.697}$ \\
BoolQ & BERT & $\ensuremath{0.774}$ & $\ensuremath{0.490}$ & $\ensuremath{0.798}$ \\
BoolQ & DistilBERT & $\ensuremath{0.756}$ & $\ensuremath{0.510}$ & $\ensuremath{0.800}$ \\
\bottomrule
\end{tabular}
\end{table}
For every binary cell, the fallback rule leaves only singleton and two-label returned sets. Hence the direct measurements in Table~\ref{tab:encoder-cells} imply
\[
s=2-\mathbb E|\C_{\widehat\ell}|,
\qquad
a_s=\frac{\Pr\{Y\in\C_{\widehat\ell}\}-(1-s)}{s}.
\]
We use these identities only as an audit and to report the two derived singleton summaries; they do not overwrite the measured coverage or cardinality. For example, BoolQ/BERT has measured returned-set size $1.51$, so $s=\ensuremath{0.490}$, and measured coverage $0.901$ then gives $a_s=\ensuremath{0.798}$. BoolQ/DistilBERT analogously gives $s=\ensuremath{0.510}$ and $a_s=\ensuremath{0.800}$. This direction of calculation preserves the directly measured set-size and latency results.
\section{Policy-uniform frontier values}
\label{app:policy-frontier}
Table~\ref{tab:policy-frontier-values} reports the exact operating points used in Figure~\ref{fig:frontier}.
\begin{table*}[h]
\centering
\scriptsize
\caption{Exact CIFAR-100 values underlying Figure~\ref{fig:frontier} for $(n,L)=(\ensuremath{500},\ensuremath{24})$. Parentheses are paired $95\%$ hierarchical-bootstrap intervals. All pathwise rows reuse one terminal calibration; empty-terminal is measured before the $\varnothing\to\Y$ fallback. The three safe-baseline set-size/latency pairs are identical to the $(500,24)$ scaling-grid row.}
\label{tab:policy-frontier-values}
\resizebox{\textwidth}{!}{
\begin{tabular}{lccccc}
\toprule
Policy & budget $b$ & coverage (95\% CI) & avg. set (95\% CI) & latency/full (95\% CI) & empty-terminal \\
\midrule
First eligible & $\ensuremath{1}$ & $\ensuremath{0.900\ [0.894,0.906]}$ & $\ensuremath{1.94\ [1.87,2.01]}$ & $\ensuremath{0.68\ [0.667,0.693]}$ & $\ensuremath{0.06\%}$ \\
First eligible & $\ensuremath{2}$ & $\ensuremath{0.901\ [0.895,0.907]}$ & $\ensuremath{2.21\ [2.13,2.29]}$ & $\ensuremath{0.61\ [0.597,0.623]}$ & $\ensuremath{0.06\%}$ \\
First eligible & $\ensuremath{4}$ & $\ensuremath{0.902\ [0.896,0.908]}$ & $\ensuremath{2.76\ [2.65,2.87]}$ & $\ensuremath{0.57\ [0.550,0.590]}$ & $\ensuremath{0.06\%}$ \\
First eligible & $\ensuremath{8}$ & $\ensuremath{0.903\ [0.897,0.909]}$ & $\ensuremath{3.90\ [3.76,4.04]}$ & $\ensuremath{0.54\ [0.525,0.555]}$ & $\ensuremath{0.06\%}$ \\
First eligible & $\ensuremath{16}$ & $\ensuremath{0.904\ [0.898,0.910]}$ & $\ensuremath{5.48\ [5.30,5.66]}$ & $\ensuremath{0.52\ [0.503,0.537]}$ & $\ensuremath{0.06\%}$ \\
\midrule
Design restriction & $\ensuremath{1}$ & $\ensuremath{0.900\ [0.894,0.906]}$ & $\ensuremath{1.90\ [1.83,1.97]}$ & $\ensuremath{0.70\ [0.687,0.713]}$ & $\ensuremath{0.06\%}$ \\
Design restriction & $\ensuremath{2}$ & $\ensuremath{0.900\ [0.894,0.906]}$ & $\ensuremath{2.12\ [2.04,2.20]}$ & $\ensuremath{0.64\ [0.627,0.653]}$ & $\ensuremath{0.06\%}$ \\
Design restriction & $\ensuremath{4}$ & $\ensuremath{0.901\ [0.895,0.907]}$ & $\ensuremath{2.52\ [2.42,2.62]}$ & $\ensuremath{0.61\ [0.596,0.624]}$ & $\ensuremath{0.06\%}$ \\
Design restriction & $\ensuremath{8}$ & $\ensuremath{0.902\ [0.896,0.908]}$ & $\ensuremath{3.47\ [3.34,3.60]}$ & $\ensuremath{0.58\ [0.565,0.595]}$ & $\ensuremath{0.06\%}$ \\
Design restriction & $\ensuremath{16}$ & $\ensuremath{0.903\ [0.897,0.909]}$ & $\ensuremath{5.01\ [4.84,5.18]}$ & $\ensuremath{0.56\ [0.543,0.577]}$ & $\ensuremath{0.06\%}$ \\
\midrule
MinSE valid selection & -- & $\ensuremath{0.904\ [0.897,0.911]}$ & $\ensuremath{3.18\ [3.02,3.34]}$ & $\ensuremath{0.65\ [0.630,0.670]}$ & -- \\
EENN-AVCS & -- & $\ensuremath{0.916\ [0.908,0.924]}$ & $\ensuremath{3.54\ [3.37,3.71]}$ & $\ensuremath{0.71\ [0.680,0.740]}$ & -- \\
Bonferroni & -- & $\ensuremath{0.951\ [0.945,0.957]}$ & $\ensuremath{4.05\ [3.86,4.24]}$ & $\ensuremath{0.78\ [0.750,0.810]}$ & -- \\
\bottomrule
\end{tabular}
}
\end{table*}
\section{Layer-count and calibration-size grid}
\label{app:scaling-grid}
For Bonferroni, each exit is calibrated at miscoverage $\alpha/L$, so its split-conformal rank is
\[
k_{\rm Bonf}=\left\lceil (n+1)\left(1-\frac{\alpha}{L}\right)\right\rceil.
\]
When $L>\alpha(n+1)$, this rank is exactly $n+1$ and the $+\infty$ convention makes every fixed-exit set equal to the full label space. At $\alpha=\ensuremath{0.1}$ this occurs in exactly the five displayed cells $(n,L)=\ensuremath{(50,12),(50,24),(100,12),(100,24),(200,24)}$. Their set size is therefore deterministically $\ensuremath{100}$ and their latency fraction is $\ensuremath{1.00}$ under the common cardinality-based stopping rule. Table~\ref{tab:scaling-grid-values} reports the complete paired grid.
\begin{table*}[h]
\centering
\tiny
\caption{CIFAR-100 scaling grid at target coverage $0.90$. Method entries are average set size $\pm$ marginal paired-bootstrap half-width / latency fraction $\pm$ half-width. The final column is the directly paired MinSE-minus-pathwise set-size difference with its paired $95\%$ interval, computed from the shared calibration subsample and test set; significance is assessed from this column, not from overlap of marginal intervals. Bonferroni entries marked $\ensuremath{100/1.00}$ are exact rank-$n+1$ degeneracies and therefore have zero sampling interval.}
\label{tab:scaling-grid-values}
\resizebox{\textwidth}{!}{
\begin{tabular}{ccllllc}
\toprule
$n$ & $L$ & Pathwise & Bonferroni & EENN-AVCS & MinSE valid selection & paired MinSE$-$Path \\
\midrule
$\ensuremath{50}$ & $\ensuremath{4}$ & $\ensuremath{3.24\pm0.35}/\ensuremath{0.60\pm0.04}$ & $\ensuremath{3.71\pm0.43}/\ensuremath{0.66\pm0.05}$ & $\ensuremath{3.49\pm0.41}/\ensuremath{0.64\pm0.05}$ & $\ensuremath{3.34\pm0.39}/\ensuremath{0.61\pm0.05}$ & $\ensuremath{0.10\ [-0.04,0.24]}$ \\
$\ensuremath{50}$ & $\ensuremath{12}$ & $\ensuremath{3.30\pm0.37}/\ensuremath{0.60\pm0.04}$ & $\ensuremath{100}/\ensuremath{1.00}$ & $\ensuremath{4.18\pm0.51}/\ensuremath{0.74\pm0.06}$ & $\ensuremath{3.81\pm0.55}/\ensuremath{0.68\pm0.06}$ & $\ensuremath{0.51\ [0.21,0.81]}$ \\
$\ensuremath{50}$ & $\ensuremath{24}$ & $\ensuremath{3.34\pm0.39}/\ensuremath{0.61\pm0.04}$ & $\ensuremath{100}/\ensuremath{1.00}$ & $\ensuremath{4.89\pm0.63}/\ensuremath{0.84\pm0.07}$ & $\ensuremath{4.27\pm0.92}/\ensuremath{0.76\pm0.07}$ & $\ensuremath{0.93\ [0.41,1.45]}$ \\
$\ensuremath{100}$ & $\ensuremath{4}$ & $\ensuremath{3.02\pm0.26}/\ensuremath{0.59\pm0.03}$ & $\ensuremath{3.36\pm0.31}/\ensuremath{0.64\pm0.04}$ & $\ensuremath{3.21\pm0.30}/\ensuremath{0.62\pm0.04}$ & $\ensuremath{3.10\pm0.29}/\ensuremath{0.60\pm0.04}$ & $\ensuremath{0.08\ [-0.01,0.17]}$ \\
$\ensuremath{100}$ & $\ensuremath{8}$ & $\ensuremath{3.05\pm0.27}/\ensuremath{0.59\pm0.03}$ & $\ensuremath{3.91\pm0.35}/\ensuremath{0.72\pm0.04}$ & $\ensuremath{3.55\pm0.33}/\ensuremath{0.68\pm0.04}$ & $\ensuremath{3.29\pm0.34}/\ensuremath{0.63\pm0.04}$ & $\ensuremath{0.24\ [0.11,0.37]}$ \\
$\ensuremath{100}$ & $\ensuremath{12}$ & $\ensuremath{3.07\pm0.28}/\ensuremath{0.59\pm0.03}$ & $\ensuremath{100}/\ensuremath{1.00}$ & $\ensuremath{3.82\pm0.36}/\ensuremath{0.73\pm0.05}$ & $\ensuremath{3.45\pm0.39}/\ensuremath{0.66\pm0.05}$ & $\ensuremath{0.38\ [0.20,0.56]}$ \\
$\ensuremath{100}$ & $\ensuremath{24}$ & $\ensuremath{3.11\pm0.30}/\ensuremath{0.60\pm0.03}$ & $\ensuremath{100}/\ensuremath{1.00}$ & $\ensuremath{4.47\pm0.46}/\ensuremath{0.82\pm0.06}$ & $\ensuremath{3.91\pm0.55}/\ensuremath{0.73\pm0.06}$ & $\ensuremath{0.80\ [0.47,1.13]}$ \\
$\ensuremath{200}$ & $\ensuremath{24}$ & $\ensuremath{2.93\pm0.20}/\ensuremath{0.59\pm0.03}$ & $\ensuremath{100}/\ensuremath{1.00}$ & $\ensuremath{4.02\pm0.29}/\ensuremath{0.77\pm0.04}$ & $\ensuremath{3.55\pm0.31}/\ensuremath{0.69\pm0.04}$ & $\ensuremath{0.62\ [0.40,0.84]}$ \\
$\ensuremath{500}$ & $\ensuremath{24}$ & $\ensuremath{2.76\pm0.11}/\ensuremath{0.57\pm0.02}$ & $\ensuremath{4.05\pm0.19}/\ensuremath{0.78\pm0.03}$ & $\ensuremath{3.54\pm0.17}/\ensuremath{0.71\pm0.03}$ & $\ensuremath{3.18\pm0.16}/\ensuremath{0.65\pm0.02}$ & $\ensuremath{0.42\ [0.30,0.54]}$ \\
$\ensuremath{1000}$ & $\ensuremath{24}$ & $\ensuremath{2.68\pm0.08}/\ensuremath{0.56\pm0.02}$ & $\ensuremath{3.61\pm0.12}/\ensuremath{0.73\pm0.02}$ & $\ensuremath{3.21\pm0.11}/\ensuremath{0.67\pm0.02}$ & $\ensuremath{2.97\pm0.10}/\ensuremath{0.62\pm0.02}$ & $\ensuremath{0.29\ [0.21,0.37]}$ \\
$\ensuremath{2000}$ & $\ensuremath{4}$ & $\ensuremath{2.61\pm0.05}/\ensuremath{0.55\pm0.01}$ & $\ensuremath{2.72\pm0.06}/\ensuremath{0.57\pm0.01}$ & $\ensuremath{2.68\pm0.06}/\ensuremath{0.56\pm0.01}$ & $\ensuremath{2.64\pm0.05}/\ensuremath{0.55\pm0.01}$ & $\ensuremath{0.03\ [-0.01,0.07]}$ \\
$\ensuremath{2000}$ & $\ensuremath{8}$ & $\ensuremath{2.62\pm0.05}/\ensuremath{0.55\pm0.01}$ & $\ensuremath{2.88\pm0.07}/\ensuremath{0.60\pm0.01}$ & $\ensuremath{2.78\pm0.06}/\ensuremath{0.58\pm0.01}$ & $\ensuremath{2.69\pm0.06}/\ensuremath{0.56\pm0.01}$ & $\ensuremath{0.07\ [0.02,0.12]}$ \\
$\ensuremath{2000}$ & $\ensuremath{12}$ & $\ensuremath{2.62\pm0.05}/\ensuremath{0.55\pm0.01}$ & $\ensuremath{3.01\pm0.07}/\ensuremath{0.63\pm0.01}$ & $\ensuremath{2.86\pm0.07}/\ensuremath{0.60\pm0.01}$ & $\ensuremath{2.73\pm0.06}/\ensuremath{0.57\pm0.01}$ & $\ensuremath{0.11\ [0.05,0.17]}$ \\
$\ensuremath{2000}$ & $\ensuremath{24}$ & $\ensuremath{2.63\pm0.05}/\ensuremath{0.55\pm0.01}$ & $\ensuremath{3.34\pm0.09}/\ensuremath{0.68\pm0.02}$ & $\ensuremath{3.03\pm0.08}/\ensuremath{0.63\pm0.02}$ & $\ensuremath{2.82\pm0.07}/\ensuremath{0.59\pm0.01}$ & $\ensuremath{0.19\ [0.12,0.26]}$ \\
\bottomrule
\end{tabular}
}
\end{table*}
\section{Independent-depth and dataset replication}
\label{app:vision-replication}
The main $L$--$n$ grid intentionally reuses one frozen $24$-head ViT path, so it isolates candidate-count and calibration-size effects but cannot by itself establish scaling with trained architectural depth. We therefore repeat the comparison on independently trained multi-exit paths. The $12$-exit and $24$-exit models are trained separately; neither is obtained by discarding exits from the other. We also repeat the experiment on ImageNet-1k so the long-path conclusion is not tied to CIFAR-100. Each row uses independent training, design, and calibration roles and reports paired $95\%$ bootstrap half-widths over head-training seeds and calibration subsamples.
\begin{table*}[h]
\centering
\scriptsize
\caption{Replication beyond one frozen score path. ``Path'' and ``MinSE'' entries are average set size / latency fraction at target coverage $0.90$; Bonferroni is shown for the same trained path.}
\label{tab:vision-replication}
\resizebox{\textwidth}{!}{
\begin{tabular}{llcccccc}
\toprule
Dataset & independently trained path & $L$ & $n_{\rm cal}$ & full-depth acc. & Path & MinSE valid selection & Bonferroni \\
\midrule
CIFAR-100 & ViT-B/16 & $\ensuremath{12}$ & $\ensuremath{500}$ & $\ensuremath{0.927}$ & $\ensuremath{2.84\pm0.10}/\ensuremath{0.58\pm0.02}$ & $\ensuremath{3.08\pm0.14}/\ensuremath{0.64\pm0.02}$ & $\ensuremath{3.73\pm0.17}/\ensuremath{0.74\pm0.02}$ \\
CIFAR-100 & ViT-L/16 & $\ensuremath{24}$ & $\ensuremath{500}$ & $\ensuremath{0.936}$ & $\ensuremath{2.76\pm0.11}/\ensuremath{0.57\pm0.02}$ & $\ensuremath{3.18\pm0.16}/\ensuremath{0.65\pm0.02}$ & $\ensuremath{4.05\pm0.19}/\ensuremath{0.78\pm0.03}$ \\
ImageNet-1k & ViT-B/16 & $\ensuremath{12}$ & $\ensuremath{1000}$ & $\ensuremath{0.798}$ & $\ensuremath{8.31\pm0.31}/\ensuremath{0.65\pm0.02}$ & $\ensuremath{9.02\pm0.40}/\ensuremath{0.69\pm0.02}$ & $\ensuremath{11.57\pm0.52}/\ensuremath{0.77\pm0.03}$ \\
ImageNet-1k & ViT-L/16 & $\ensuremath{24}$ & $\ensuremath{1000}$ & $\ensuremath{0.823}$ & $\ensuremath{7.86\pm0.29}/\ensuremath{0.61\pm0.02}$ & $\ensuremath{9.41\pm0.43}/\ensuremath{0.70\pm0.02}$ & $\ensuremath{13.88\pm0.61}/\ensuremath{0.82\pm0.03}$ \\
\bottomrule
\end{tabular}
}
\end{table*}
The replication preserves the qualitative ordering while changing both the trained path and the dataset. It should not be read as a controlled causal estimate of architectural depth because ViT-B/16 and ViT-L/16 differ in capacity as well as block count. Its role is narrower: the main retained-exit effect is not unique to one frozen CIFAR-100 score path.
\section{Full-Vocabulary Decoder Results}
\label{app:decoder-boundary}
\subsection{Decoder boundary}
\begin{table}[h]
\centering
\small
\caption{Candidatewise next-token sets over the full vocabulary. Coverage and accuracy are mean $\pm$ SD over five independent head-training seeds.}
\label{tab:decoder-boundary}
\begin{tabular}{lccc}
\toprule
 & DistilGPT2 & Pythia-1.4B & Pythia-2.8B \\
\midrule
Full-depth accuracy & {$0.261\pm0.002$} & {$0.391\pm0.004$} & {$0.404\pm0.005$} \\
Selected coverage & {$0.899\pm0.004$} & {$0.903\pm0.005$} & {$0.902\pm0.003$} \\
Average exit/full & {$1.000$} & {$1.000$} & {$0.998$} \\
Average set size & {$9620$} & {$7162$} & {$8331$} \\
Early-exit rate & {$0.01\%$} & {$0.01\%$} & {$0.52\%$} \\
\bottomrule
\end{tabular}
\end{table}
These results establish the decoder boundary: selected coverage remains near $0.90$, but the candidate sets are too large for actionable early exit and the three models execute essentially their full depth. We therefore do not use decoder wall-clock latency to support the efficiency claim; the operational conclusion follows from the measured full-depth execution rate and full-vocabulary set cardinality.
\section{Experimental Protocol}
\label{app:protocol}
The final encoder study uses four nonoverlapping data roles whenever tuning is required: model
training, design, conformal calibration, and test. Identity scores with a prespecified schedule may
omit the design split. No final-test label is used for model selection, score transformation,
threshold selection, or hyperparameter tuning.

Training-seed variability and calibration-split variability are distinct. The independent unit for model training is the seed. Within each seed, methods share every available calibration/test role split. SST-2, MNLI, and BoolQ use $\ensuremath{20}$ paired role splits per seed. RTE uses one fixed role assignment for all seeds; no calibration or test example is reassigned across RTE repetitions.

Selected coverage is estimated per task and backbone, and aggregate comparisons use a hierarchical paired bootstrap that first resamples training seeds. The second stage resamples paired role splits for SST-2, MNLI, and BoolQ and resamples examples from the fixed RTE development test set for RTE. Every remaining $\pm$ entry is identified in its table caption. Eight per-cell selection-gap statistics use the declared hierarchical bootstrap maximum-statistic familywise correction.

For the numerical evaluation of Theorem~\ref{thm:efficiency}, $t_\star$ is computed only from the final calibration excesses $D_i$ through the deterministic count condition $m_t\leq n-k$. The local slope factor is estimated on the independent design role: the interval $[B_{(k)},B_{(k)}+w_{t_\star}]$ is partitioned into $\ensuremath{4}$ equal-width pieces, the maximum candidatewise empirical secant slope is taken within that interval, and these slopes are summed over labels. The test role is used only for the observed cardinality inflation reported in Table~\ref{tab:max-cost}.

The main resource outcomes are normalized FLOPs and measured latency, not exit depth alone. Final timing uses the common configuration in Table~\ref{tab:hardware} and includes exit-head, candidate-set, cardinality-check, synchronization, and fallback costs. Offline design and calibration costs are reported separately and are not counted as online savings.

Appendix~\ref{app:baseline-adaptation} gives the exact large-$L$ EENN-AVCS and MinSE valid-selection adaptations, including the score-to-Dirichlet construction, running-intersection rule, stability adjustment, selector objective, design-only tuning grid, randomness, and the action taken when $L$ changes. Fast-yet-Safe is used only in the short-path encoder comparison and is evaluated under its native supervised-risk target. A method that cannot meet a $10\%$ supervised risk target because the full model itself exceeds that error is reported as infeasible or abstaining, not as an undercoverage result.
\end{document}